\UseRawInputEncoding
\documentclass[10.5pt,twocolumn,journal,letterpaper,review]{IEEEtran}

\pdfoutput=1 

\usepackage{amsmath, amsthm, amssymb}
\usepackage{url,flushend,multirow,booktabs}
\usepackage[ruled,linesnumbered]{algorithm2e}  % for cross line vertically
\usepackage{graphicx}
\usepackage{bm}
\usepackage{cite}
\usepackage{subfigure}
\usepackage{float}
\usepackage{color}
\usepackage{multirow}
\usepackage[colorlinks,citecolor=green,urlcolor=blue,bookmarks=false,hypertexnames=true]{hyperref}
\usepackage{colortbl}
\definecolor{maroon}{cmyk}{0.08,0.04,0.00,0.06}

\newcommand{\eg}{e.g.}
\newcommand{\ie}{i.e.}

\newcommand{\etal}{\textit{et al.}}

\DeclareFixedFont{\mf}{OT1}{ptm}{m}{n}{10pt}
\DeclareFixedFont{\mfb}{OT1}{ptm}{bx}{n}{10pt}

\begin{document}
%

\title{Fast Fourier Inception Networks for Occluded Video Prediction}

\author{Ping~Li,~\IEEEmembership{Member,~IEEE}, Chenhan~Zhang, Xianghua~Xu
\thanks{P.~Li, C.~Zhang, and X.~Xu are with the School of Computer Science and Technology, Hangzhou Dianzi University, Hangzhou 310018, China, and P.~Li is also with the Guangdong Laboratory of Artificial Intelligence and Digital Economy (SZ), Shenzhen 518132, China  (e-mail: patriclouis.lee@gmail.com, zch2020@hdu.edu.cn, xhxu@hdu.edu.cn).}
}

%\markboth{IEEE TRANSACTIONS ON NEURAL NETWORKS AND LEARNING SYSTEMS, ~Vol.~x, No.~x, xx~2017}
%{LI \MakeLowercase{\textit{et al.}}:~ONLINE ROBUST LOW-RANK TENSOR MODELING FOR STREAMING DATA ANALYSIS}
\markboth{Draft}
{LI \MakeLowercase{\textit{et al.}}:~Fast Fourier Inception Networks for Occluded Video Prediction}

\maketitle

\begin{abstract}
  Video prediction is a pixel-level task that generates future frames by employing the historical frames. There often exist continuous complex motions, such as object overlapping and scene occlusion in video, which poses great challenges to this task. Previous works either fail to well capture the long-term temporal dynamics or do not handle the occlusion masks. To address these issues, we develop the fully convolutional Fast Fourier Inception Networks for video prediction, termed \textit{FFINet}, which includes two primary components, \ie, the occlusion inpainter and the spatiotemporal translator. The former adopts the fast Fourier convolutions to enlarge the receptive field, such that the missing areas (occlusion) with complex geometric structures are filled by the inpainter. The latter employs the stacked Fourier transform inception module to learn the temporal evolution by group convolutions and the spatial movement by channel-wise Fourier convolutions, which captures both the local and the global spatiotemporal features. This encourages generating more realistic and high-quality future frames. To optimize the model, the recovery loss is imposed to the objective, \ie, minimizing the mean square error between the ground-truth frame and the recovery frame. Both quantitative and qualitative experimental results on five benchmarks, including Moving MNIST, TaxiBJ, Human3.6M, Caltech Pedestrian, and KTH, have demonstrated the superiority of the proposed approach. Our code is available at \href{https://github.com/mlvccn/research/tree/main/FFINet_VidPre}{GitHub}.  
   
\end{abstract}


 \ifCLASSOPTIONpeerreview
 \begin{center} \bfseries EDICS Category: 3-BBND \end{center}
 \fi

\IEEEpeerreviewmaketitle

\section{Introduction}
\label{sec1:intro}

\IEEEPARstart{V}{ideo} prediction is the pixel-level task of predicting future frames given past video frames. It has great potential in real-world applications, \eg, climate forecast \cite{shi-nips2015-convlstm}\cite{wu-cvpr2021-motionrnn}, autonomous driving \cite{bei-cvpr2021-sadm}\cite{castrejon-iccv2019-conditionalvrnn}, and robot control \cite{babaeizadeh-iclr2018-sv2p}\cite{denton-icml2018-svg}. From the randomness perspective, the video prediction methods are divided into two categories, \ie, \textit{deterministic} and \textit{stochastic}. Given a video, the former assumes that the future is deterministic and yields the single prediction, while the latter assumes that there are multiple predictions. However, the latter requires large computations, which hinders its wide applications in highly-demanding scenarios. Hence, this work mainly focuses on the deterministic video prediction.

Most previous works either adopt Recurrent Neural Networks (RNNs) \cite{elman-cognsci1990-rnn} or Convolutional Neural Networks (CNNs) \cite{lecun-proc1998-cnn} to predict future frames. Typically, Predictive RNN (PreRNN) \cite{wang-nips2017-predrnn}, Eidetic 3D Long-Short Term Memory (E3D-LSTM) \cite{wang-iclr2019-e3dlstm}, and Motion-Aware Unit (MAU) \cite{chang-nips2021-mau} adopt RNNs to capture the temporal dynamics; Deep Voxel Flow (DVF) \cite{liu-iccv2017-dvf}, Disentangling Propagation \& Generation (DPG) \cite{gao-iccv2019-dpg}, and Simple Video Prediction (SimVP) \cite{gao-cvpr2022-simvp} use CNNs to capture the inter-frame dependency. In addition, Vision Transformer (ViT) \cite{dosovitskiy-iclr2021-vit} has been proved ineffective in video prediction \cite{gao-cvpr2022-simvp}, which is because ViT requires large-scale data for training but the training videos in video prediction task are usually insufficient.
%such as Temporal Convolutional Transformer Network (TCTN) \cite{yang-arxiv2021-tctn}.
  
When there exist heavy overlaps among multiple moving objects in video, it is usually difficult to completely recover the object appearances and requires more past frames without overlap. To address this issue, E3D-LSTM \cite{wang-iclr2019-e3dlstm} and CrevNet (Conditionally Reversible Network) \cite{yu-iclr2020-crevnet} use 3D convolution layers to enlarge the temporal receptive field with larger kernel sizes, which are computationally expensive; 3D convolutions are also used in video saliency prediction \cite{wang-tmm2023-stsanet}. Besides, MAU \cite{chang-nips2021-mau} computes the attention map between the current and the past features for capturing long-term dependency. However, these methods adopt the costly RNNs which desire long training time, \eg, it requires 10 cycles for predicting 10 frames and can not use parallelization. This inspires us to employ the Fast Fourier Transform (FFT) \cite{chi-nips2020-ffc} to capture the global spatiotemporal receptive field by designing the \textbf{FFT Inception} module. It employs the group convolutions to model the local temporal dynamics and uses the channel-wise fast Fourier transform \cite{katznelson-amm2005-harmonic} to capture the global motion trend. 

Moreover, previous video prediction methods assume the frames are clean, failing to consider the frames with occlusions, possibly caused by the camera pollution or the object occlusion. To tackle this challenge, one should first inpaint the occluded area and then predict the future frames. Particularly, we adopt the Fast Fourier Convolution (FFC) \cite{suvorov-wacv2022-lama} module to recover the missing area by an \textbf{inpainter}, and impose the recovery loss (\ie, MSE-Mean Square Error) onto the inpainter to minimize the error between the recovery frame and the source clean frame. 

\begin{figure*}[!t]
 	\centering
 	\includegraphics[width=0.8\linewidth]{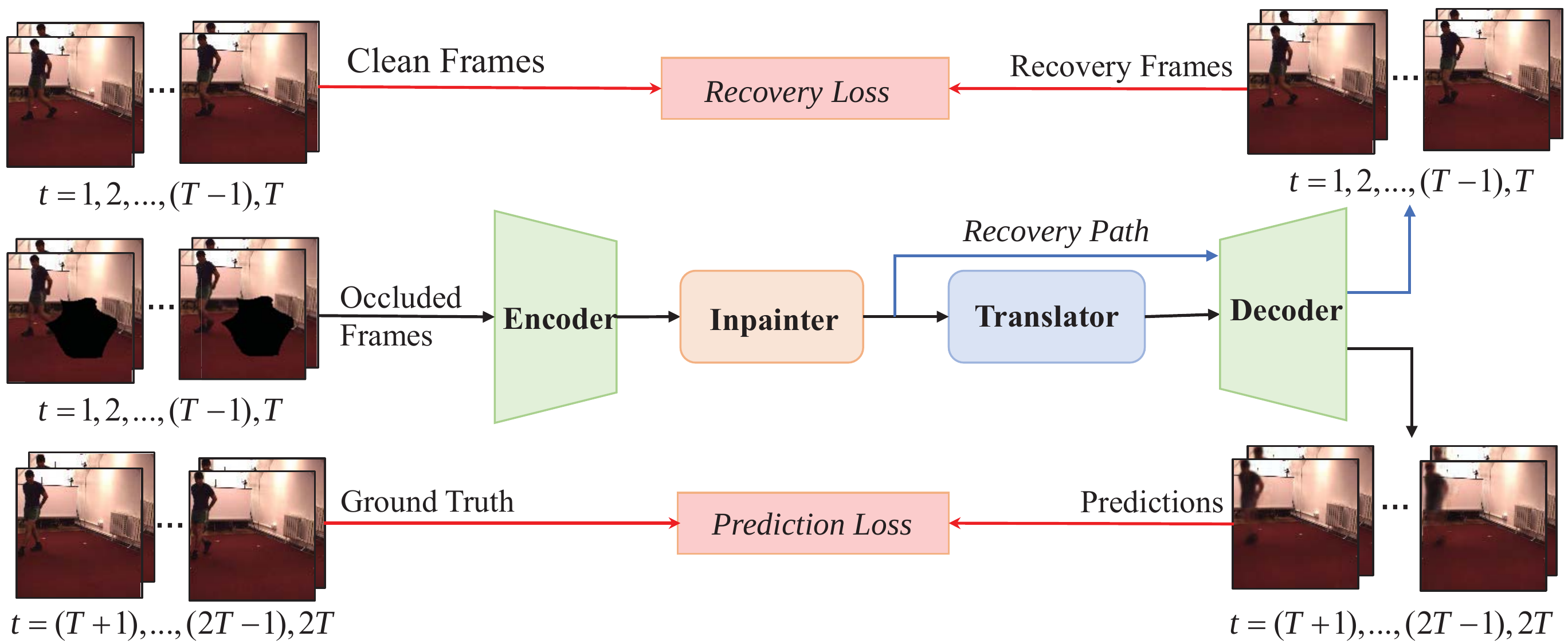}
 	\caption{Illustration of the occluded video prediction task. Given $T$ input frames, it predicts the future $T$ frames. Both the recovery loss and the prediction loss adopt the Mean Squared Error (MSE).}
 	\label{fig:illustration}
 	%\vspace{-1mm}
\end{figure*}

Hence, we propose the fully-convolutional \textbf{F}ast \textbf{F}ourier \textbf{I}nception \textbf{Net}works (\textbf{FFINet}) for video prediction with occlusion, as illustrated in Fig.~\ref{fig:illustration}. It is composed of an encoder, an inpainter, a translator, and a decoder. Here, the \textbf{translator} is stacked with a series of FFT Inception modules, which are used for capturing both the local and global spatiotemporal features. 
 
% ------------------------- Illustration of Occluded Video Prediction -------------------
 
Our main contributions are highlighted in the following:
\begin{itemize}
  \item To our best, we are the first to explore the occluded video prediction, and propose the fully-convolutional fast Fourier Inception Networks (FFINet) for this task.
  
  \item Our framework includes an inpainter that consists of the Fast Fourier Convolution modules to recover the missing areas in video, and a translator that consists of the Fast Fourier Transform inception modules to learn the spatiotemporal features for predicting future frames.
    
  \item Empirical studies on several benchmarks including Moving MNIST \cite{srivastava-icml2015-movingmnist}, TaxiBJ \cite{zhang-aaai2017-trafficbj}, Human3.6M \cite{ionescu--tpami2014-human3.6m}, Caltech Pedestrian \cite{geiger-ijrr2013-kitti}, and KTH \cite{schuldt-icpr2004-kth}, have demonstrated the superiority of the proposed FFINet framework for video prediction with occlusions.

\end{itemize}

The rest of this paper is organized as follows. Section~\ref{related} reviews some closely related works and Section~\ref{method} describes the proposed FFINet framework. Then, experimental results are reported in Section~\ref{test} with rigorous analysis. Finally, we conclude this work in Section~\ref{conclusion}.
%

%-------------------------------------------------------------------------
\section{Related Work}
\label{related}
This section mainly discusses several encoder-translator-decoder architectures of video prediction, including full RNN, RNN or ViT (as translator) with CNN, and full CNN.

\subsection{Full RNN}
The full-RNN methods employ the stacked RNNs for encoding, spatiotemporal feature translation, and decoding. For example, Convolutional LSTM \cite{shi-nips2015-convlstm} extends the fully-connected LSTMs to the architecture with convolutions, and the memory updates only along the temporal dimension; PredRNN \cite{wang-nips2017-predrnn} uses the spatiotemporal memory flow to update the memory states along both the spatial and the temporal dimensions, but it suffers from the gradient propagation difficulty when capturing the long-term dependency, which is addressed by PredRNN++ \cite{wang-icml2018-predrnn++}. Moreover, MIM (Memory In Memory) \cite{wang-cvpr2019-mim} models both the non-stationary and stationary properties by self-updated memory; to reduce the computational and memory costs, fRNN (folded RNN) \cite{oliu-eccv2018-frnn} shares the state between the encoder and decoder layers, while the representation is stratified during learning; Conv-TT-LSTM \cite{su-nips2020-convttlstm} (Convolutional Tensor-Train LSTM) extends ConvLSTM from the first-order to the higher-order scenario by combining convolutional features across time, \eg, update the current features by employing the past continuous features. In addition, Motion RNN \cite{wu-cvpr2021-motionrnn} divides the physical motion into the transient variation and motion trend, which are updated simultaneously; PredRNNv2 \cite{wang-tpami2023-predrnnv2} develops a decoupling loss and a reverse scheduled sampling strategy to extend the original PredRNN.  

\subsection{RNN or ViT with CNN}
This type uses RNN or ViT as the translator, and CNN as the encoder and decoder. CNN is used to learn the spatial feature and RNN or ViT is used to model temporal dynamics. For example, E3D-LSTM \cite{wang-iclr2019-e3dlstm} uses 3D convolution to extract spatiotemporal features, while CrevNet \cite{yu-iclr2020-crevnet} includes the flow-based encoder and decoder to capture information-preserving features. But they fail to predict the motion trend during more complex scenarios. So, PhyDNet (Physical Dynamics Network) \cite{guen-cvpr2020-phydnet} and Vid-ODE (Video generation with Ordinary Differential Equation) \cite{park-aaai2021-vidode} use the partial-differential equation or ODE to enhance the physical dynamics modeling; DDME (Dynamic Motion Estimation and Evolution) \cite{kim-tmm2021-dmee} employs the dynamic convolution to generate convolution kernel at each moment to estimate the temporal dynamics; some works use the dynamic convolution kernel in the fully convolutional network to generate the talking face video \cite{ye-tmm2023-dck}. The above methods are good at capturing the short-term dependency, but the predicted frames become obscure when the time gets longer. To capture the long-term dependency, LMC (Long-term Motion Context) \cite{lee-cvpr2021-lmc} uses a memory to save the long-term motion context, which is used for predicting the future; MAU \cite{chang-nips2021-mau} captures the inter-frame motion by broadening the temporal receptive field of the predictive units. 

Furthermore, some attempts have been devoted to the sub-tasks in video prediction. For example, Chang \etal~\cite{chang-cvpr2022-strpm} have developed the spatiotemporal residual predictive model for high-resolution video prediction by focusing more on frame details; CPL (Continual Predictive Learning) \cite{chen-cvpr2022-cpl} presents the mixture world model and the predictive experience replay strategy to alleviate the continual learning problem; MAC (Modular Action Concept network) \cite{yu-cvpr2022-mac} considers the semantic action-conditional video prediction, which predicts future frames according to the semantic labels that describe the action interactions. 

In addition, Vision Transformer (ViT) \cite{dosovitskiy-iclr2021-vit} has been used for modeling the latent dynamics, \eg, Weissenborn \etal~\cite{weissenborn-iclr2020-subscalevitr} design a simple auto-regressive video generation model with a 3D self-attention mechanism to yield continuous frames; Rakhimov \etal~\cite{rakhimov-visigrapp2021-latentvit} directly uses ViT to model the latent dynamics; TCTN (Temporal Convolutional Transformer Network) \cite{yang-arxiv2021-tctn} employs the transformer-based encoder with temporal convolution layers to capture both the short-term and long-term dependencies. However, ViT model requires large-scale training data to achieve satisfying performance, and it does not bring about much gains on video prediction since the training data in this task is usually insufficient.

\subsection{Full CNN}
This type fully uses CNN to learn and update the spatiotemporal features. Some works will additionally use optical flow to assist the video prediction, such as DVF (Deep Voxel Flow) \cite{liu-iccv2017-dvf} employs the CNN auto-encoder to learn the voxel flow to reconstruct the frame by using nearby frame voxel flow; Wu \etal~\cite{wu-cvpr2022-ovp} treat video prediction as the video frame interpolation optimization, which is affected by the optical flow quality; DPG (Disentangling Propagation and Generation) \cite{gao-iccv2019-dpg} and LCVG (Layered Controllable Video Generation) \cite{huang-eccv2022-lcvp} both separate the foreground from the background, where the former uses optical flow and the latter uses CNN for discrimination. Moreover, Generative Adversarial Network (GAN) \cite{goodfellow-nips2014-gan} is applied to video prediction for increasing the authenticity and the continuity of predicted frames, \eg, rCycleGAN (Retrospective Cycle GAN) \cite{kwon-cvpr2019-rcyclegan} enhances the temporal consistency by learning cycle GAN; Xu \etal~\cite{xu-ijcv2021-pma} develop a progressive multiple granularity analysis framework to match the prototype motion dynamics with the input sequence. However, the above methods desire complex modules and training skills to improve the performance. Recently, SimVP \cite{gao-cvpr2022-simvp} is a simple video prediction model using common convolution modules, and achieves the State-Of-The-Art (SOTA) performance. But its temporal receptive field is still narrow, limiting its performance upgrade. This inspires us to design an efficient module to enlarge the temporal receptive field to capture the long-term motion dynamics. 

% ------------------------- Fast Fourier Inception Network Framework for Occluded Video Prediction -------------------
\begin{figure*}[!t]
	\centering
	\includegraphics[width=0.95\linewidth]{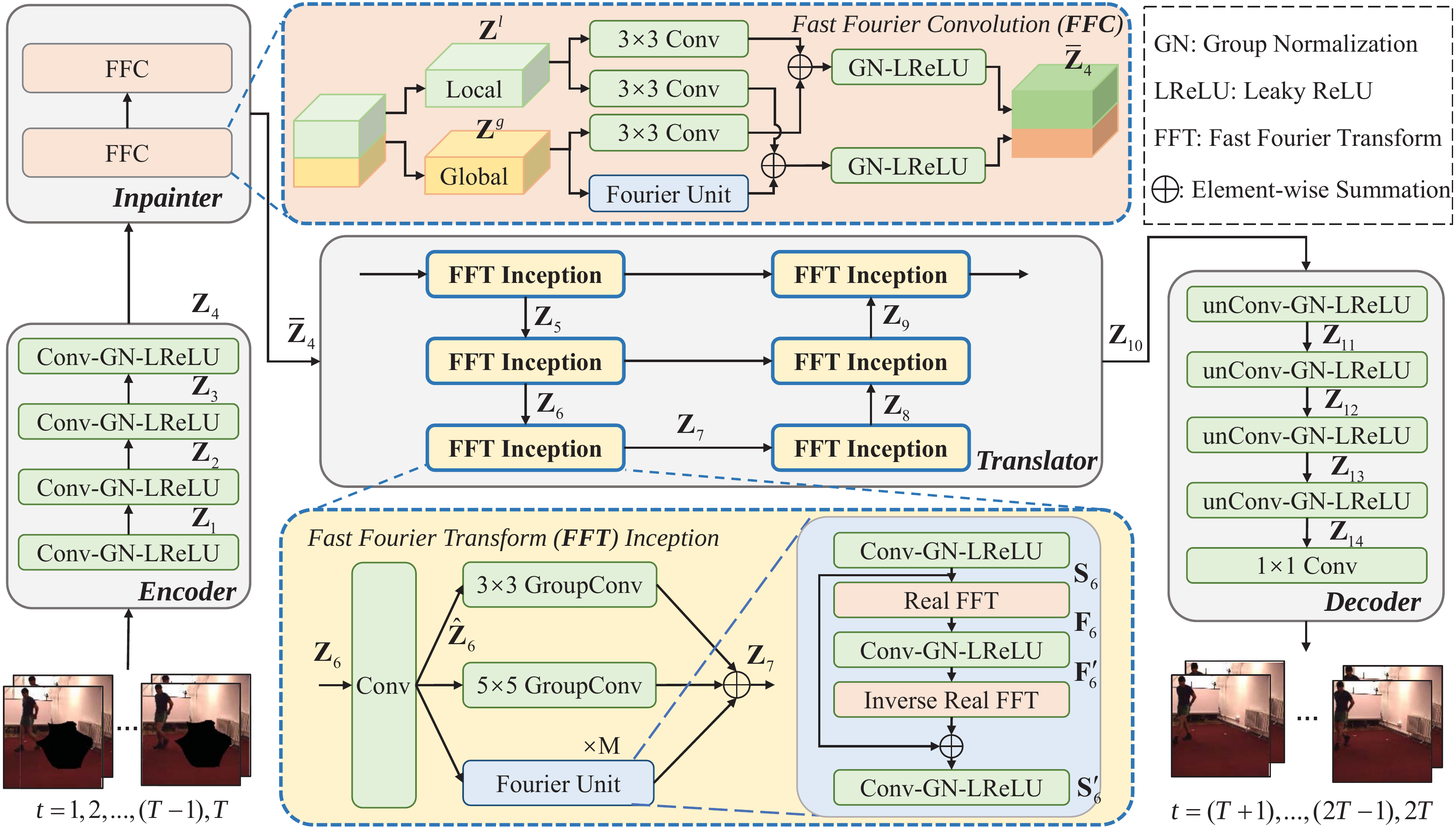}
	\caption{Framework of the Fast Fourier Inception Networks (FFINet) for occluded video prediction. Here, $(\tilde{N}, \hat{N})$=(4,6), $M$ is the number of Fourier Units.}
	\label{fig:framework}
	%\vspace{-1mm}
\end{figure*}

% ------------
\section{Our FFINet Method}
\label{method}
This section mainly describes the proposed FFINet framework as depicted in Fig.~\ref{fig:framework}, which includes encoder, inpainter, translator, and decoder.

\subsection{Problem Formulation}
Given a video sequence with $T$ frames, \ie, $\mathcal{X}_T=\{\mathbf{X}_t\in \mathbb{R}^{C\times H\times W}|t=1,2, \ldots, T\}$, where $\mathbf{X}_t$ denotes the $t$-th RGB frame with width $W$, height $H$, and $C$ channels, the video prediction task aims to generate the future $T^\prime$ frames, \ie,  $\mathcal{Y}_{T^\prime}=\{\mathbf{X}_t \}_{t = T+1}^{T+T^\prime}$, by learning a mapping function $\mathcal{F}_\theta : \mathcal{X}_T \mapsto \mathcal{Y}_{T^\prime}$, where $\theta$ is the model parameter. The vanilla goal is to minimize the loss between the ground-truth frames $\mathcal{Y}_{T^\prime}$ and the predicted frames $\mathcal{F}_\theta(\mathcal{X}_T)$. For occluded video prediction, the model should first inpaint the corrupted frames and then predict the future frames. Given the occluded video sequence $\hat{\mathcal{X}}_T = \{\hat{\mathbf{X}}_t\in \mathbb{R}^{C\times H\times W}|t=1,2, \ldots, T\}$, the goal is to minimize the loss between the ground-truth frames $\mathcal{Y}_{T^\prime}$ and the predicted frames $\mathcal{F}^\prime_\theta(\hat{\mathcal{X}}_T)$ with inpainting. 

\subsection{Encoder}
The encoder is used to learn spatial features of video frames, and consists of $\tilde{N}$ stacked $3\times 3$ Conv2D blocks with group normalization ${\rm GN}(\cdot)$ to speed up the convergence and leaky ReLU function $\sigma(\cdot)$ to enhance the feature nonlinearity. Mathematically, the hidden feature is computed by 
\begin{equation}
	\label{eq:encoder}
	\mathbf{Z}_i = \sigma({\rm GN}({\rm Conv2D}(\mathbf{Z}_{i-1}))),1\leq i \leq \tilde{N},
\end{equation}
where $\mathbf{Z}_0 \in \mathbb{R}^{(B\cdot T)\times C\times H\times W}$ is the input video sequence. Here $B$ denotes the batch size, and we do the downsampling every two convolution layers by halving the height and the width of the feature map. At the last convolution layer, we concatenate the spatial features along the temporal dimension and reshape them to the tensor $\mathbf{Z}_{\tilde{N}} \in \mathbb{R}^{B\times (T\cdot \tilde{C})\times H^\prime \times W^\prime}$, where $\tilde{C}$, $H^\prime$, $W^\prime$ are the feature channel, height, and width. Generally, the feature dimension is reduced, and thus the computational cost can be saved a lot. The obtained features will be fed into the inpainter for recovering the missing areas.

\subsection{Inpainter}
The inpainter is used to recover the occluded frames and it adopts two Fast Fourier Convolution (FFC) \cite{suvorov-wacv2022-lama} modules. Each FFC module consists of two inter-connected branches, \ie, common convolution layer on the half of the feature channels, and channel-wise fast Fourier transform on the rest channels. The former captures the local spatial features and the latter captures the global context. 

The input of the inpainter is the encoded feature $\mathbf{Z}_{\tilde{N}}$, and we learn the local feature $\mathbf{Z}^l \in \mathbb{R}^{B \times \frac{T \cdot \widetilde{C}}{2} \times H^{\prime} \times W^{\prime}}$ according to 
\begin{equation}
	\label{eq:local}
	\mathbf{Z}^l = \sigma({\rm GN}({\rm Conv2D}(\mathbf{Z}^l) + {\rm Conv2D}(\mathbf{Z}^g))),
\end{equation}
where ${\rm Conv2D}$ adopts the kernel size of $3\times 3$. The first convolution layer is used to obtain the local feature, and the second convolution layer is used to fuse both the local and the global features.

Similarly, we learn the global feature $\mathbf{Z}^g \in \mathbb{R}^{B\times \frac{T \cdot \tilde{C}}{2} \times H^{\prime} \times W^{\prime}}$ according to
\begin{equation}
	\label{eq:global}
	\mathbf{Z}^g = \sigma({\rm GN}({\rm Conv2D}(\mathbf{Z}^l) + {\rm FU}(\mathbf{Z}^g))),
\end{equation}
where the $3\times 3$ Conv2D fuses the local and the global features, and ${\rm FU(\cdot)}$ denotes the Fourier Unit that includes the channel-wise fast Fourier transform and $1\times 1$ Conv2D, which has the global receptive field.
The fused features from the two branches are concatenated along the channel, leading to the recovery feature tensor $\bar{\mathbf{Z}}_{\tilde{N}}\in \mathbb{R}^{B \times (T \cdot \tilde{C}) \times H^{\prime} \times W^{\prime}}$.

\subsection{Translator}
The translator learns the temporal evolution by capturing and updating the spatiotemporal features. Inspired by \cite{gao-cvpr2022-simvp}, it is composed of $\hat{N}$ stacked Fast Fourier Transform (FFT) Inception blocks ${\rm FFTI(\cdot)}$ (See the middle bottom in Fig.~\ref{fig:framework}), and the hidden feature $\mathbf{Z}_j\in \mathbb{R}^{B \times \hat{C} \times H^{\prime} \times W^{\prime}}$ is obtained by
\begin{equation}
	\label{eq:translator}
	\mathbf{Z}_j = {\rm FFTI}(\mathbf{Z}_{j-1}),\tilde{N} < j \leq \tilde{N} + \hat{N},
\end{equation}
where $\hat{C}$ is the channel number, which keeps still across succeeding FFT Inception blocks in the translator. FFT Inception block mainly involves the convolution layers with the kernel sizes of $1\times 1$, $3\times 3$, and $5\times 5$, as well as the Fourier Units. Here, $1\times 1$ Conv2D reduces the feature dimension from $T\cdot \tilde{C}$ (the first block) or $\hat{C}$ (the remaining blocks) to $\frac{\hat{C}}{2}$, while the $3\times 3$ and $5\times 5$ Conv2D adopt the group convolutions, \ie, equally dividing the feature channels into 8 groups and each group captures distinct local pattern of the features. For the last block, the output channel is recovered to $T^\prime \cdot \tilde{C}$.

Note that our FFT Inception block abandons the large convolution kernels to save the computational costs, and introduces the Fourier Unit ${\rm FU}(\cdot)$ to capture the global context in early layers. The Fourier Unit (FU) has a receptive field covering the entire frame by including the channel-wise fast Fourier transform and $1\times 1$ Conv2D at lower cost. In particular, FU is composed of three convolution layers with group normalization and leaky ReLU function, real FFT, and inverse real FFT (${\rm IFFT}(\cdot)$) layers. The convolution layers capture the local context and the real FFT models the global context. The first and the third convolution layers are used to align the feature dimension, while the second convolution layer is used to update the spatiotemporal features in the frequency domain. Here, the real FFT (using half of the FFT spectrum) is applied to real-valued spatial features and the inverse real FFT makes the recovered spatial feature be real valued \cite{chi-nips2020-ffc}. The details can be expressed by
\begin{equation}
	\label{eq:S_j}
	\mathbf{S}_j = \sigma({\rm GN}({\rm Conv2D}(\hat{\mathbf{Z}}_j)))\in \mathbb{R}^{B\times \frac{\hat{C}}{2} \times H^{\prime}\times W^{\prime}}, 
\end{equation}
\begin{equation}
	\label{eq:F_j}
	\mathbf{F}_j = {\rm Real\ FFT}(\mathbf{S}_j) \in \mathbb{C}^{B\times \hat{C} \times H^\prime\times \frac{W^\prime}{2}}, 
\end{equation}
\begin{equation}
	\label{eq:F_j_prime}
	\mathbf{F}_j^{\prime} = \sigma({\rm GN}({\rm Conv2D}(\mathbf{F}_j))) \in \mathbb{C}^{B\times \hat{C} \times H^\prime\times \frac{W^\prime}{2}},   
\end{equation}
\begin{equation}
	\label{eq:S_j_prime}
	\mathbf{S}_j^{\prime} = \sigma({\rm GN}({\rm Conv2D}({\rm IFFT}({\mathbf{F}}_j^{\prime})+ \mathbf{S}_j))) \mathbb{R}^{B\times \hat{C} \times H^\prime\times W^\prime},
\end{equation}
where $\mathbb{C}$ denotes the frequency domain, and $\hat{\mathbf{Z}}_j={\rm Conv2D}(\mathbf{Z}_j)\in \mathbb{R}^{B\times \frac{\hat{C}}{2} \times H^{\prime}\times W^{\prime}}$. Here, $\mathbf{S}_j$ denotes the spatiotemporal feature in source domain, $\mathbf{F}_j$ denotes the spatiotemporal feature in frequency domain, $\mathbf{S}_j^{\prime}$ and $\mathbf{F}_j^{\prime}$ denote the updated features. In practice, we use multiple Fourier Units (\eg, $M$=3) in the model.

\subsection{Decoder}
The decoder is composed of $\tilde{N}$ stacked unConv blocks, which are used to decode the updated spatiotemporal features into future frames. Following \cite{gao-cvpr2022-simvp}, we use ConvTranspose2d to serve as the ${\rm unConv(\cdot)}$ operator for upsampling the features along the spatial dimension. The hidden feature can be computed by
\begin{equation}
	\label{eq:decoder}
	\mathbf{Z}_k = \sigma({\rm GN}({\rm unConv}(\mathbf{Z}_{k-1}))),\tilde{N}+ \hat{N} < k \leq 2\tilde{N} + \hat{N},
\end{equation}
where $k$ indexes the unConv block. 

Given the updated spatiotemporal feature $\mathbf{Z}_{\tilde{N}+ \hat{N}}\in \mathbb{R}^{B\times (T^\prime \cdot \tilde{C}) \times H^{\prime}\times W^{\prime}}$, the decoder outputs the future $T^\prime$ frames $\mathcal{Y}_{T^{\prime}}\in \mathbb{R}^{B\times T^{\prime}\times C\times H\times W}$.

\subsection{Loss Function}
There are two losses in our FFINet model, including the prediction loss $\mathcal{L}_{pre}$ and the recovery loss $\mathcal{L}_{rec}$, both of which adopt the MSE function, \ie,
\begin{equation}
	\label{eq:loss_total}
	\mathcal{L} = \mathcal{L}_{pre} + \lambda \mathcal{L}_{rec},
\end{equation}
where the constant $\lambda>0$ governs the contribution of the inpainter to the objective. It becomes the traditional video prediction when the inpainter is removed.

The prediction loss minimizes the error between the ground-truth frames $\mathcal{Y}_T^\prime = \{\mathbf{X}_t\}_{t=T+1}^{T+T^\prime}$  and the predicted frames $\mathcal{F}_\theta(\mathcal{X}_T)$, \ie, 
\begin{equation}
	\label{eq:loss_pre}
	\mathcal{L}_{pre} = \min\limits_{\theta}\sum_{t=T+1}^{T+T^\prime}\left \| \mathbf{X}_t-\mathcal{F}_{\theta}(\mathcal{X}_T)_t \right \|^2_2,
\end{equation}
where $\mathbf{X}_t$ is the $t$-th ground-truth frame, $\mathcal{F}_{\theta}(\mathcal{X}_T)_t$ is the $t$-th predicted frame, $\mathcal{X}_T$ denotes the training frames, $\theta$ is the model parameter, and $\|\cdot\|$ denotes the $\ell_2$-norm. 

The recovery loss minimizes the error between the ground-truth frames $\mathcal{Y}_T = \{\mathbf{X}_t\}_{t=1}^T$ and the recovery frames $\mathcal{R}_\phi(\hat{\mathcal{X}}_T)$ by the inpainter, \ie,
\begin{equation}
	\label{eq:loss_rec}
	\mathcal{L}_{rec} = \min\limits_{\phi}\sum_{t=1}^{T} \| \mathbf{X}_t-\mathcal{R}_{\phi}(\hat{\mathbf{X}}_t)\|^2_2,
\end{equation}
where $\phi$ is the parameter of the model without the translator.

\section{Experiments}
\label{test}
This section shows extensive experimental results on several benchmark data sets. All experiments were conducted on a machine with three NVIDIA RTX 3090 Graphics Cards, and our model was compiled using PyTorch 1.12, Python 3.10, and CUDA 11.1.

%-------------------------------------------------------------------------
\subsection{Data Sets}
In total, there are five publicly available video databases used in the experiments. Details are shown below.

\textbf{Moving MNIST}\footnote{\url{https://www.cs.toronto.edu/~nitish/unsupervised_video/}}~\cite{srivastava-icml2015-movingmnist}. It consists of paired evolving hand-written digits from the MNIST\footnote{\url{http://yann.lecun.com/exdb/mnist/}} data set. Following \cite{wang-nips2017-predrnn}, the training set includes 10000 sequences and the test set includes 5000 sequences. Each sequence consists of 20 successive $64\times 64$ frames with 2 randomly appearing digits. Among them, 10 frames are the input and the rest are the output. The initial position and rate of each digit are random, but the rate keeps the same across the entire sequence.

\textbf{TaxiBJ}\footnote{\url{https://github.com/TolicWang/DeepST/tree/master/data/TaxiBJ}}~\cite{zhang-aaai2017-trafficbj} is collected from the real-world traffic scenario in Beijing, ranging from 2013 to 2016. The traffic flows have strong temporal dependency among nearby area, and the data pre-processing follows \cite{wang-cvpr2019-mim}. The data of the last four weeks are used as the test set (1334 clips) while the rest are the training set (19627 clips). Each clip has 8 frames, where 4 frames are the input and the others are the output. The size of each video frame is $32\times 32\times 2$, and the two channels indicate the in and out traffic flow. 

\textbf{Human3.6M}\footnote{\url{http://vision.imar.ro/human3.6m/description.php}}~\cite{ionescu--tpami2014-human3.6m} contains the sports videos of 11 subjects in 17 scenes, involving 3.6 million human pose images from 4 distinct camera views. Following \cite{wang-cvpr2019-mim}, we use the data in the walking scene, which includes $128\times 128\times 3$ RGB frames. The subsets $\{S1, S5{\rm -}S8\}$ are for training (2624 clips) and $\{S9, S11\}$ are for test (1135 clips). Each clip has 8 frames, and the half of them are the input. 

\textbf{KITTI\&Caltech Pedestrian}\footnote{\url{https://www.cvlibs.net/datasets/kitti/}}. Following \cite{wang-cvpr2019-mim}, we use 2042 clips in KITTI \cite{geiger-ijrr2013-kitti} for training and 1983 clips in Caltech Pedestrian \cite{dollar-cvpr2009-caltech} for test. Both of them are driving databases taken from a vehicle in an urban environment, and the RGB frames are resized to $128\times 160$ by center-cropping and downsampling. The former includes ``city'', ``residential'', and ``road'' categories, while the latter has about 10 hours of $640\times 480$ video. Each clip has 20 consecutive frames, where 10 frames are the input and the others are the output.

\textbf{KTH}\footnote{\url{https://www.csc.kth.se/cvap/actions/}}~\cite{schuldt-icpr2004-kth} includes six action classes, \ie, walking, jogging, running, boxing, hand waving, and hand clapping, involving 25 subjects in four different scenes. Each video clip is taken in 25 fps and is 4 seconds on average. Following \cite{villegas-iclr2017-mcnet}, the gray-scale frames are resized to $128\times 128$. The training set has 5200 clips (16 subjects) and the test set has 3167 clips (9 subjects). Each clip has 30 frames, where 10 frames are the input and 20 frames are the output.

\subsection{Evaluation Metrics}
Following \cite{guen-cvpr2020-phydnet}\cite{yu-iclr2020-crevnet}\cite{gao-cvpr2022-simvp}, we employ MSE (Mean Square Error), MAE (Mean Absolute Error), SSIM (Structure Similarity Index Measure) \cite{wang-tip2004-ssim}, and PSNR (Peak Signal to Noise Ratio) to evaluate the quality of the predicted frames. On Caltech Pedestrian \cite{dollar-cvpr2009-caltech}, we use MSE, SSIM, and PSNR; on KTH \cite{schuldt-icpr2004-kth}, we use SSIM and PSNR; on the remaining ones, we use MAE, MSE, and SSIM. SSIM ranges from -1 to 1, and the images are more similar when it approaches 1. The larger the PSNR db value, the better quality it achieves.

%Given two images $\{\mathbf{x}, \mathbf{y}\} \in \mathbb{R}^{w\times h}$, their SSIM is calculated by
%\begin{equation}
%	SSIM(\mathbf{x}, \mathbf{y})=\frac{(2\mu_x\mu_y+c_1)(2\sigma_{xy}+c_2)}{(\mu_x^2+\mu_y^2)(\sigma_x^2+\sigma_y^2+c_2)}
%\end{equation}
%where $\mu$ is the average of all pixel values in the matrix, $\sigma^2$ is the matrix variance, $\sigma_{xy}$ is the covariance of the image pair $(\mathbf{x}, \mathbf{y})$, and the constants $c_1$ and $c_2$ are used to maintain the stability. SSIM ranges from -1 to 1, and the images are more similar when it approaches 1.
%
%PSNR is computed as 
%\begin{equation}
%	PSNR(\mathbf{x}, \mathbf{y})=10\cdot \log_{10}\frac{{(2^n-1)}^2}{MSE(\mathbf{x}, \mathbf{y})},
%\end{equation}
%where $MSE(\mathbf{x}, \mathbf{y})$ is the mean square error between the two images. The larger the PSNR db value, the better quality it achieves.

% ------------- Experimental settings -----------
\begin{table}[!t]
	\centering
	\caption{Experimental settings.}
	\label{tbl:setting}
	\setlength{\tabcolsep}{0.9mm}{ 
	\begin{tabular}{l c c c c c c c c}
		\toprule[0.75pt]
		Dataset                                            & $(H,W,C)$   & In$\rightarrow$Out           &$\widetilde{C}$ &$\hat{C}$ &$\widetilde{N}$ &$\hat{N}$ &$M$ &Epoch \\
		\midrule[0.5pt]
		Moving MNIST\cite{srivastava-icml2015-movingmnist} & (64,64,1)   & 10$\rightarrow$10       & 64  & 512 &  4  &  6   & 3  & 2000\\
		TaxiBJ\cite{zhang-aaai2017-trafficbj}              & (32,32,2)   & 4$\rightarrow$4         & 64  & 256 &  3  &  4   & 2 & 80\\
		Human3.6M\cite{ionescu--tpami2014-human3.6m}       & (128,128,3) & 4$\rightarrow$4         & 64  & 64  &  1  &  10  & 2 & 100 \\
		KITTI\&Caltech\cite{geiger-ijrr2013-kitti}         & (128,160,3) & 10$\rightarrow$1        & 64  & 128 &  1  &  6   & 2 & 50\\
		KTH\cite{schuldt-icpr2004-kth}                     & (128,128,1) & 10$\rightarrow$20/40 & 32  & 128 &  3  &  8   & 1 & 100 \\
		\toprule[0.75pt]
	\end{tabular}
}
\end{table}

% -------- Occluded example --------
\begin{figure}[!t]
	\centering
	\includegraphics[width=0.45\textwidth]{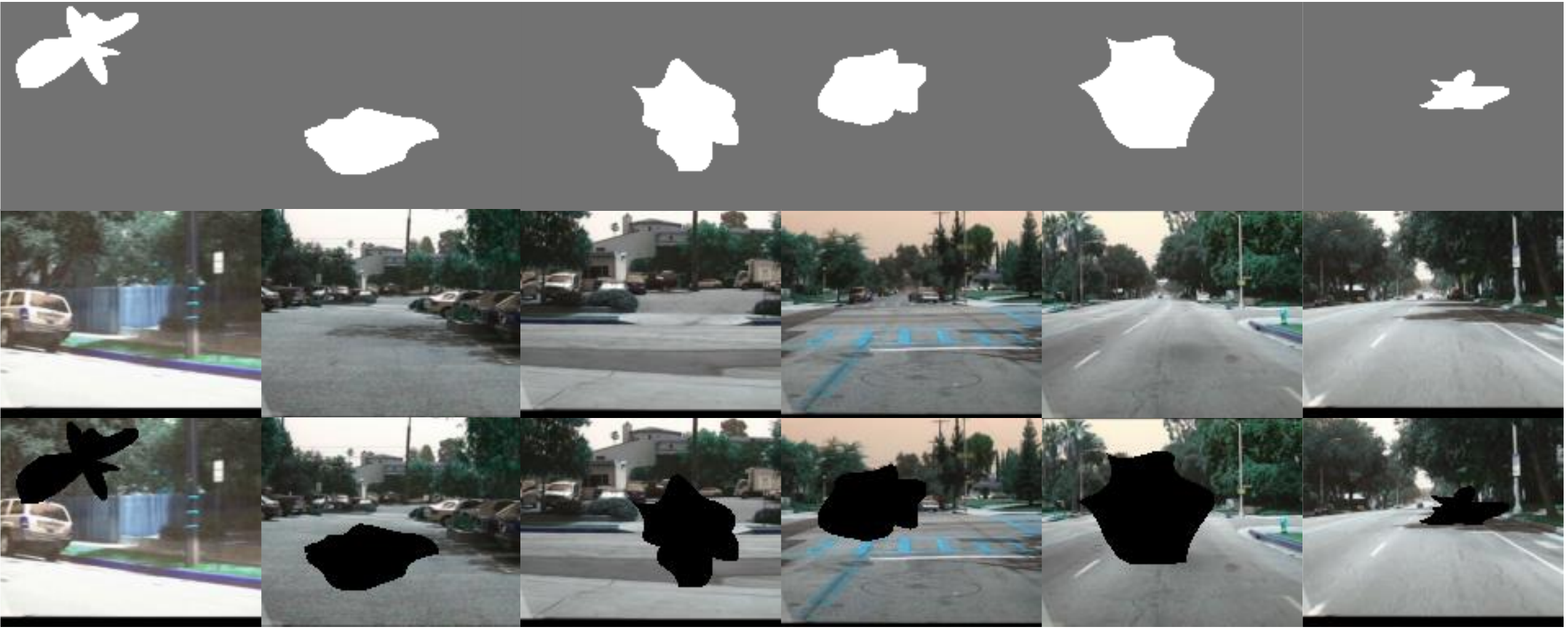}
	\caption{Occluded examples of the Caltech Pedestrian dataset. Top row: masks, middle row: source frame, bottom row: occluded frame.}
	\label{fig:mask_example}
\end{figure}

% --------- Results on Moving MNIST, TaxiBJ, and Huamn3.6M (without occlusion) -------
\begin{table*}[!t]
	\centering
	\caption{Comparison results on Moving MNIST \cite{srivastava-icml2015-movingmnist}, TaxiBJ \cite{zhang-aaai2017-trafficbj}, and Human3.6M \cite{ionescu--tpami2014-human3.6m}.}
	\label{tbl:mnist_taxibj_human}
	\small
	\setlength{\tabcolsep}{1.4mm}{
		\begin{tabular}{lc ccc c lll c lll}
			\toprule[0.75pt]
			\multirow{2}{*}{Method} & \multirow{2}{*}{Venue} & \multicolumn{4}{c}{Moving MNIST\cite{srivastava-icml2015-movingmnist}} & \multicolumn{4}{c}{TaxiBJ\cite{zhang-aaai2017-trafficbj}} & \multicolumn{3}{c}{Human3.6M\cite{ionescu--tpami2014-human3.6m}}  \\ \cmidrule[0.5pt]{3-5}  \cmidrule[0.5pt]{7-9} \cmidrule[0.5pt]{11-13}
			& & MSE$\downarrow$ & MAE$\downarrow$ & SSIM$\uparrow$ & & MSE$\downarrow$ & MAE$\downarrow$ & SSIM$\uparrow$ && MSE$\downarrow$ & MAE$\downarrow$ &SSIM$\uparrow$  \\ \midrule[0.5pt]
			PredRNN\cite{wang-nips2017-predrnn}    & NeurIPS'17& 56.8 & 126.1 & 0.867 & & 46.4 & 17.1 & 0.971 && 48.4 & 18.9 & 0.781 \\
			PredRNN++\cite{wang-icml2018-predrnn++}& ICML'18   & 46.5 & 106.8 & 0.898 & & 44.8 & 16.9 & 0.977 && 45.8 & 17.2 & 0.851 \\
			MIM\cite{wang-cvpr2019-mim}            & CVPR'19   & 44.2 & 101.1 & 0.910 & & 42.9 & 16.6 & 0.971 && 42.9 & 17.8 & 0.790 \\
			E3D-LSTM\cite{wang-iclr2019-e3dlstm}   & ICLR'19   & 41.3 & 86.4  & 0.910 & & 43.2 & 16.9 & 0.979 && 46.4 & 16.6 & 0.869 \\
			\midrule[0.5pt]
			PhyDNet\cite{guen-cvpr2020-phydnet}    & CVPR'20   & 24.4 & 70.3  & 0.947 & & 41.9 & \textbf{16.2} & \textbf{0.982} && 36.9 & 16.2 & 0.901 \\
			CrevNet\cite{yu-iclr2020-crevnet}      & ICLR'20   & 22.3 & - & \underline{0.949} & & - & - & - && - & - & - \\
			MAU\cite{chang-nips2021-mau}            & NeurIPS'21& 27.6 & 80.3  & 0.937 & &42.2$^{\ast}$  & 16.4$^{\ast}$ & 0.982$^{\ast}$ && \underline{31.2}$^{\ast}$  & 15.0$^{\ast}$& 0.885$^{\ast}$ \\
			STAM\cite{chang-tmm2022-stam}          & TMM'23    & 28.6 &   -   & 0.935 & & 44.1 &  -   &  -    &&  -  & \underline{13.2} & 0.875 \\
			SimVP\cite{gao-cvpr2022-simvp}         & CVPR'22   & 23.8 & \underline{69.9} & 0.948 & & \underline{41.4} & \textbf{16.2} & \textbf{0.982} && 31.6 & 15.1 & \underline{0.904} \\
			PredRNNv2\cite{wang-tpami2023-predrnnv2}& TPAMI'23   & \underline{19.9} &  - & 0.939 & & 45.6$^{\ast}$ & 16.8$^{\ast}$ & 0.980$^{\ast}$ && 36.3$^{\ast}$ & 17.7$^{\ast}$ & 0.863$^{\ast}$ \\   
			\midrule[0.5pt]
			Ours                                   &           & \textbf{19.2} & \textbf{60.4} & \textbf{0.958} & & \textbf{41.2} & \textbf{16.2} & \textbf{0.982} && \textbf{23.3} & \textbf{11.9} & \textbf{0.913} \\
			\toprule[0.75pt]
		\end{tabular}
	}
\end{table*}

\subsection{Experimental Setup}
\textbf{Training Phase}.
The FFINet model is trained on the training set of each database, and use Adam \cite{kingma-iclr2015-adam} optimizer with the OneCycle \cite{simth-ispp2019-onecycle} learning rate scheduler and the momentum $(\beta_1, \beta_2)$=(0.9, 0.999). The initial learning rate is 0.01, and the batch size $B$ is 16. The constant $\lambda$ is set to 0.5 for Moving MNIST \cite{srivastava-icml2015-movingmnist} and 1.0 for the rest. The other experimental settings are listed in Table~\ref{tbl:setting}, where $\tilde{N}$ and $\hat{N}$ denotes the convolution layer number and the FFT Inception block number of the encoder and the translator, respectively; $M$ is the number of Fourier Units in the FFT Inception block. Note that in the ablation study, we use 10$\rightarrow$20 for KTH dataset.

The feature height and width $(H^\prime, W^\prime)$=$(\frac{H}{2^{\lfloor\tilde{N}/2\rfloor}}, \frac{W}{2^{\lfloor\tilde{N}/2\rfloor}})$, where $\lfloor \cdot \rfloor$ is the floor function. To generate the masks, we adopt the algorithm in \cite{li-cvpr2022-e2fgvi}, which randomly produces a set of control points around a unit circle and smoothly connects them to a closed cyclic contour by cubic Bezier curves. We show some occluded examples of Caltech Pedestrian in Fig.~\ref{fig:mask_example}. The masks are randomly and instantly generated during training for enhancing the model robustness to different occlusions.

\textbf{Inference Phase}.
The masks are generated in advance and keep still during frame prediction for fairness. 
%

% ------- Results on Caltech Pedestrian (without occlusion) -----------
\begin{table}[!t]
	\centering
	\caption{Comparison results on Caltech Pedestrian \cite{dollar-cvpr2009-caltech}.}
	\label{tbl:caltech}
	\small
	\setlength{\tabcolsep}{0.95mm}{
		\begin{tabular}{l c c c c}
			\toprule[0.75pt]
			\multirow{2}{*}{Method} & \multirow{2}{*}{Venue} & \multicolumn{3}{c}{Caltech Pedestrian\cite{dollar-cvpr2009-caltech}(10$\rightarrow$1) } \\ \cmidrule[0.5pt]{3-5} 
			& &MSE$\downarrow$ & SSIM$\uparrow$ & PSNR$\uparrow$(db) \\
			\midrule[0.5pt]
			DVF\cite{liu-iccv2017-dvf}                  & ICCV'17   & -    & 0.897 & 26.2 \\
			Dual-GAN\cite{liang-iccv2017-dualmotiongan} & ICCV'17   & 2.41 & 0.899 & -    \\
			PredNet\cite{lotter-iclr2017-prednet}       & ICLR'17   & 2.42 & 0.905 & 27.6 \\
			CtrlGen\cite{hao-cvpr2018-ctrlgen}          & CVPR'18   & -    & 0.900 & 26.5 \\
			ContextVP\cite{byeon-eccv2018-contextvp}    & ECCV'18   & 1.94 & 0.921 & 28.7 \\
			DPG\cite{gao-iccv2019-dpg}                  & ICCV'19   & -    & 0.923 & 28.2 \\
			STMFANet\cite{jin-cvpr2020-stmfanet}        & CVPR'20   & 1.59 & 0.927 & 29.1 \\
			CrevNet\cite{yu-iclr2020-crevnet}           & ICLR'20   & 1.55 & 0.925 & 29.3 \\
			MAU\cite{chang-nips2021-mau}                & NeurIPS'21& 1.24 & 0.943 & \underline{30.1} \\
			VPCL\cite{geng-cvpr2022-vpcl}               & CVPR'22   & -    & 0.928 & -    \\
			SimVP\cite{gao-cvpr2022-simvp}              & CVPR'22   & 1.56 & 0.940 & \textbf{33.1} \\
			STAM\cite{chang-tmm2022-stam}                & TMM'23   & \textbf{1.11} & \underline{0.945} & 29.9 \\
			\midrule[0.5pt]
			Ours &  & \underline{1.14} & \textbf{0.949} & \textbf{33.1} \\
			\toprule[0.75pt]
		\end{tabular}
	}
\end{table}

% ------- Results on KTH (without occlusion) -----------
\begin{table}[!t]
	\centering
	\caption{Comparison results on KTH \cite{schuldt-icpr2004-kth}.}
	\label{tbl:kth}
	\small
	\setlength{\tabcolsep}{0.85mm}{
		\begin{tabular}{l c c c c c}
			\toprule[0.75pt]
			\multirow{2}{*}{Method} & \multirow{2}{*}{Venue} & \multicolumn{2}{c}{KTH\cite{schuldt-icpr2004-kth}(10$\rightarrow$20)} & \multicolumn{2}{c}{KTH\cite{schuldt-icpr2004-kth}(10$\rightarrow$40)}\\ \cmidrule[0.5pt]{3-6} 
			& &SSIM$\uparrow$ & PSNR$\uparrow$(db) & SSIM$\uparrow$ & PSNR$\uparrow$(db) \\
			\midrule[0.5pt]
			%ConvLSTM\cite{shi-nips2015-convlstm}   & \scriptsize{NeurIPS'15} & 0.712 & 23.58 & 0.639 & 22.85 \\
			DFN\cite{jia-nips2016-dfn}             & \scriptsize{NeurIPS'16} & 0.794 & 27.26 & 0.652 & 23.01 \\
			MCnet\cite{villegas-iclr2017-mcnet}    & ICLR'17    & 0.804 & 25.95 & 0.730 & 23.89 \\
			PredRNN\cite{wang-nips2017-predrnn}    & \scriptsize{NeurIPS'17} & 0.839 & 27.55 & 0.703 & 24.16 \\
			fRNN\cite{oliu-eccv2018-frnn}          & ECCV'18    & 0.771 & 26.12 & 0.678 & 23.77 \\
			SV2P\cite{babaeizadeh-iclr2018-sv2p}   & ICLR'18    & 0.838 & 27.79 & 0.789 & 26.12 \\
			\scriptsize{PredRNN++}\cite{wang-icml2018-predrnn++}& ICML'18    & 0.865 & 28.47 & 0.741 & 25.21 \\
			SAVP\cite{lee-iclr2019-savp}           & ICLR'19    & 0.852 & 27.77 & 0.811 & 26.18 \\
			\scriptsize{E3D-LSTM}\cite{wang-iclr2019-e3dlstm}   & ICLR'19    & 0.879 & 29.31 & 0.810 & 27.24 \\
			\scriptsize{STMFANet}\cite{jin-cvpr2020-stmfanet}   & CVPR'20    & 0.893 & 29.85 & 0.851 & 27.56 \\
			GridVP\cite{gao-iros2021-gridkeypoint} & IROS'21    & -     & -     & 0.837 & 27.11 \\
			SimVP\cite{gao-cvpr2022-simvp}         & CVPR'22    & \underline{0.905} & \underline{33.72} & \underline{0.886} & \underline{32.93} \\
			\scriptsize{PredRNNv2}\cite{wang-tpami2023-predrnnv2}& TPAMI'23   & 0.838 & 28.37 &   -   &   -   \\
			\midrule[0.5pt]
			Ours &  & \textbf{0.912} & \textbf{34.24} & \textbf{0.894} & \textbf{33.28} \\
			\toprule[0.75pt]
		\end{tabular}
	}
\end{table}

% ------------ Computations on Moving MNIST -----------
\begin{table}[!t]
	\centering
	\caption{Computation comparison on Moving MNIST \cite{srivastava-icml2015-movingmnist}.}
	\label{tbl:computation}
	\small
	\setlength{\tabcolsep}{0.95mm}{
		\begin{tabular}{l r  r r c c c}
			\toprule[0.75pt]
			\multirow{2}{*}{Method}  & \multirow{2}{*}{Venue}  & FLOPs & Train & Test &\#Params   & MSE \\
			&   & (G)$\downarrow$ & (s)$\downarrow$ & (fps)$\uparrow$ & (M)$\downarrow$   &  $\downarrow$ \\
			\midrule[0.5pt]
			PredRNN\cite{wang-nips2017-predrnn}     & \scriptsize{NeurIPS}'17 & 115.6  & 300   & 107  & 23.8  &56.8 \\
			\scriptsize{PredRNN++}\cite{wang-icml2018-predrnn++} & ICML'18 & 171.7  & 530   & 70   & 38.6  &46.5 \\
			MIM\cite{wang-cvpr2019-mim}             & CVPR'19 & 179.2  & 564   & 55   & 38.0  &44.2 \\
			\scriptsize{E3D-LSTM}\cite{wang-iclr2019-e3dlstm}    & ICLR'19 & 298.9  & 1417  & 59   & 51.3  &41.3 \\
			\midrule[0.5pt]
			CrevNet\cite{yu-iclr2020-crevnet}       & ICLR'20 & 270.7  & 1030  & 10   & ~5.0   &22.3 \\
			PhyDNet\cite{guen-cvpr2020-phydnet}     & CVPR'20 & 15.3   & 196   & 63   & ~\textbf{3.1} &24.4 \\
			MAU\cite{chang-nips2021-mau}            & \scriptsize{NeurIPS}'21 & 17.8   & 210   & 58   & ~\underline{4.5}   &27.6 \\
			SimVP\cite{gao-cvpr2022-simvp}          & CVPR'22 & 19.4   & 86    & 190  &  22.3 &23.8 \\
			\midrule[0.5pt]
			%Ours($M$=2) &   & \textbf{7.12}  & \textbf{66} &  \textbf{303}  & 14.5 & \underline{22.1} \\
			Ours($M$=3) &   & \textbf{7.83}  & \textbf{81} & \textbf{204} & 19.1 & \textbf{19.2} \\
			\toprule[0.75pt]
		\end{tabular}
	}
\end{table}

\subsection{Quantitative Results}
We show the quantitative comparison results in two situations, including video prediction with and without occlusion. On Moving MNIST \cite{srivastava-icml2015-movingmnist}, TaxiBJ \cite{zhang-aaai2017-trafficbj}, and Human3.6M \cite{ionescu--tpami2014-human3.6m}, we compare PredRNN \cite{wang-nips2017-predrnn}, PredRNN++ \cite{wang-icml2018-predrnn++}, PredRNNv2 \cite{wang-tpami2023-predrnnv2}, MIM (Memory In Memory) \cite{wang-cvpr2019-mim}, E3D-LSTM \cite{wang-iclr2019-e3dlstm}, PhyDNet (Physical Dynamics Network) \cite{guen-cvpr2020-phydnet}, MAU (Motion-Aware Unit) \cite{chang-nips2021-mau}, CrevNet \cite{yu-iclr2020-crevnet}, and SimVP \cite{gao-cvpr2022-simvp}; on Caltech Pedestrian \cite{dollar-cvpr2009-caltech}, we compare PredNet \cite{lotter-iclr2017-prednet}, ContextVP \cite{byeon-eccv2018-contextvp}, STMFANet (Spatial-Temporal Multi-Frequency Analysis Network) \cite{jin-cvpr2020-stmfanet}, CrevNet (Conditionally Reversible Network) \cite{yu-iclr2020-crevnet}, Dual-GAN (Dual Generative Adversarial Network) \cite{liang-iccv2017-dualmotiongan}, DVF (Deep Voxel Flow) \cite{liu-iccv2017-dvf}, CtrlGen (Controllable video Generation) \cite{hao-cvpr2018-ctrlgen}, DPG (Disentangling Propagation and Generation) \cite{gao-iccv2019-dpg}, VPCL (Video Prediction with Correspondence-wise Loss) \cite{geng-cvpr2022-vpcl}, and SimVP \cite{gao-cvpr2022-simvp}; on KTH \cite{schuldt-icpr2004-kth}, we compare PredRNN \cite{wang-nips2017-predrnn}, PredRNN++ \cite{wang-icml2018-predrnn++}, fRNN \cite{oliu-eccv2018-frnn}, MCNet \cite{villegas-iclr2017-mcnet}, E3D-LSTM \cite{wang-iclr2019-e3dlstm}, SV2P (Stochastic Variational Video Prediction) \cite{babaeizadeh-iclr2018-sv2p}, STMFANet \cite{jin-cvpr2020-stmfanet}, GridVP \cite{gao-iros2021-gridkeypoint}, DFN (Dynamic Filter Network) \cite{jia-nips2016-dfn}, SAVP (Stochastic Adversarial Video Prediction) \cite{lee-iclr2019-savp}, and SimVP \cite{gao-cvpr2022-simvp}. The best records are highlighted in bold and the second-best ones are underlined; ``-'' indicates the record is unavailable; ``*'' indicates the record is obtained by re-implementing the code provided by the authors. 

\textbf{Traditional video prediction}.
The performance comparison results on Moving MNIST \cite{srivastava-icml2015-movingmnist}, TaxiBJ \cite{zhang-aaai2017-trafficbj}, and Human3.6M \cite{ionescu--tpami2014-human3.6m} are shown in Table~\ref{tbl:mnist_taxibj_human}. Following \cite{gao-cvpr2022-simvp}, MSE values are enlarged by 100 times for TaxiBJ, MSE and MAE values are divided by 10 and 100 for Human3.6M. Note that the methods in the top group adopt the fixed training set, and the rest generate the training sample online, \eg, randomly select two digits and their motion path. From the table, it can be seen that our method consistently outperforms the other competitive alternatives across all evaluation metrics. Compared to the strongest baseline, our FFINet model reduces the MAE by 9.5 on Moving MNIST, while it reduces MSE by 7.9 on Human3.6M. Previous methods like PredRNN \cite{wang-nips2017-predrnn}, PredRNN++ \cite{wang-icml2018-predrnn++}, MIM \cite{wang-cvpr2019-mim} use recurrent neural networks to model the temporal dynamics, failing to capture the long-term dependency and thus obtaining the poor predictions. E3D-LSTM \cite{wang-iclr2019-e3dlstm} and CrevNet \cite{yu-iclr2020-crevnet} adopt the 3D convolution to enlarge the receptive field but largely increase the computational costs. By contrast, our method employs the Fast Fourier Transform Inception blocks as the translator to better capture the spatiotemporal tendency by learning both the local and the global spatiotemporal features in video, resulting in higher-quality predictions.

Moreover, the performance comparison results on Caltech Pedestrian \cite{dollar-cvpr2009-caltech} and KTH \cite{schuldt-icpr2004-kth} are shown in Table~\ref{tbl:caltech} and Table~\ref{tbl:kth}, respectively. From the tables, we observe that our approach has the most satisfying overall performance on predicting future frames in comparison with several SOTA methods. This demonstrates that the stacked Fourier transform inception blocks are able to learn the temporal evolution by adopting group convolutions and the channel-wise Fourier convolutions. 

In addition, we show the efficiency comparison results on Moving MNIST in Table~\ref{tbl:computation}, where the training time is computed for one epoch (per frame) using a single RTX3090 and the test time is the average fps of 10,000 samples. As seen from the table, our FFINet method enjoys the best prediction performance at the lowest training time with the fast inference speed. For example, our FLOPS is less than the half of the best candidate SimVP \cite{gao-cvpr2022-simvp} but with much lower MSE.  

\textbf{Occluded video prediction}.
The results of occluded video prediction are shown in Table~\ref{tbl:occlusion_mnist}, Table~\ref{tbl:occlusion_taxibj}, Table~\ref{tbl:occlusion_human}, Table~\ref{tbl:occlusion_caltech}, and Table~\ref{tbl:occlusion_kth}, for Moving MNIST \cite{srivastava-icml2015-movingmnist}, TaxiBJ \cite{zhang-aaai2017-trafficbj}, Human3.6M \cite{ionescu--tpami2014-human3.6m}, Caltech Pedestrian \cite{dollar-cvpr2009-caltech}, and KTH \cite{schuldt-icpr2004-kth}, respectively. Here, ``Occ.'' denotes whether the video is occluded by the masks. %Note that the frame resolution in TaxiBJ \cite{zhang-aaai2017-trafficbj} dataset is too low and it becomes very difficult to predict frames when adding the occlusion, so the results are not reported.  

From these tables, we can see that the video prediction performance degenerates a lot when the video frames are occluded by random masks, which indicates the occlusions really do harm to predicting future frames. Moreover, when we use the Fast Fourier Convolution blocks to build the inpainter for recovering the missing areas in the video frames, the video prediction performance is improved, \eg, the MAE value reduces from 68.4 to 65.8 on Moving MNIST. This demonstrates that it is beneficial for the model to fill the missing areas by employing the inpainter to enlarge the receptive field. Note that the performance is slightly boosted with occlusion for some methods like MAU \cite{chang-nips2021-mau} and PredRNNv2 \cite{wang-tpami2023-predrnnv2} on TaxiBJ \cite{zhang-aaai2017-trafficbj}, which might be the reason that there are many repeated patterns in the traffic flow and the masks are treated as noise to increase the model robustness.

% ------------- Results on Moving MNIST (with occlusion) --------------
\begin{table}[!t]
	\centering
	\caption{Comparison results on Moving MNIST \cite{srivastava-icml2015-movingmnist} with occlusion.}
	\label{tbl:occlusion_mnist}
	\small
	\setlength{\tabcolsep}{0.85mm}{
	\begin{tabular}{l c c l l l}
		\toprule[0.75pt]
		Method & Venue & Occ. & MSE$\downarrow$ & MAE$\downarrow$ & SSIM$\uparrow$  \\
		\midrule[0.5pt]
		\rowcolor{maroon}CrevNet\cite{yu-iclr2020-crevnet}       & ICLR'20    &            & 30.2       & 86.3                   & 0.935         \\
		&            & \checkmark & 35.0(+4.8) & 94.2(\underline{+7.9}) & 0.915 \\
		\rowcolor{maroon}PhyDNet\cite{guen-cvpr2020-phydnet}     & CVPR'20    &            & 24.4       & 70.3                   & 0.947         \\
		&            & \checkmark & 29.1(+4.7) & 80.0(+9.7)             & 0.936 \\
		\rowcolor{maroon}MAU\cite{chang-nips2021-mau}            & NeurIPS'21 &            & 27.6       &80.3                    & 0.937         \\
		&            & \checkmark & 35.7(+8.1) &98.6(+18.3)             & 0.913 \\
		\rowcolor{maroon}SimVP\cite{gao-cvpr2022-simvp}          & CVPR'22    &            & 23.8       & 68.9                   & 0.948         \\
		&            & \checkmark & 28.7(+5.1) & 80.8(+11.9)            & 0.936 \\
		\rowcolor{maroon}\scriptsize{PredRNNv2}\cite{wang-tpami2023-predrnnv2}& TPAMI'23   &            & 27.4       &82.2                    & 0.937         \\
		&            & \checkmark & 32.8(+5.4) & 92.3(+10.1)            & 0.925 \\
		\midrule[0.5pt]
		\rowcolor{maroon}Ours                                    &            &            & 19.2       & 60.4                   & 0.958         \\
		Ours               & w/o Inpainter   & \checkmark & \underline{22.7}(+3.5)& \underline{68.4}(+8.0)& \underline{0.950} \\
		Ours                  & w/ Inpainter       & \checkmark & \textbf{21.7}(+2.5)& \textbf{65.8}(+5.4)& \textbf{0.952}\\
		\toprule[0.75pt]
	\end{tabular}
}
\end{table}

% ------------- Results on TaxiBJ (with occlusion) --------------
\begin{table}[!t]
	\centering
	\caption{Comparison results on TaxiBJ \cite{zhang-aaai2017-trafficbj} with occlusion.}
	\label{tbl:occlusion_taxibj}
	\small
	\setlength{\tabcolsep}{0.85mm}{
	\begin{tabular}{l c c l l l}
		\toprule[0.75pt]
		Method & Venue & Occ. & MSE$\downarrow$ & MAE$\downarrow$ & SSIM$\uparrow$  \\
		\midrule[0.5pt]
		\rowcolor{maroon}PhyDNet\cite{guen-cvpr2020-phydnet}     & CVPR'20    &            & 41.9       & 16.2                   & 0.982         \\
		&            & \checkmark & 43.0(+1.1) & 16.6(+0.4)             & 0.981 \\
		\rowcolor{maroon}MAU\cite{chang-nips2021-mau}            & NeurIPS'21 &            & 42.2       & 16.4                   & 0.982         \\
		&            & \checkmark & 41.7(-0.5) &16.3(\underline{-0.1})             & 0.982 \\
		\rowcolor{maroon}SimVP\cite{gao-cvpr2022-simvp}          & CVPR'22    &            & 41.4       & 16.2                   & 0.982         \\
		&            & \checkmark & 43.3(+1.9) & 16.8(+0.6)            & 0.981 \\
		\rowcolor{maroon}PredRNNv2\cite{wang-tpami2023-predrnnv2}& TPAMI'23   &            & 45.6       & 16.8                   & 0.980         \\
		&            & \checkmark & 45.2(-0.4) & 16.8(+0.0)            & 0.980\\
		\rowcolor{maroon}Ours                                    &            &            & 41.2       & 16.2                   & 0.982         \\
		Ours   & w/o Inpainter     & \checkmark & \underline{40.6}(-0.6)& \underline{16.1}(-0.1) & \textbf{0.983} \\
		Ours   &  w/ Inpainter      & \checkmark & \textbf{40.4}(-0.8)  & \textbf{16.0}(-0.2) & \textbf{0.983} \\
		\toprule[0.75pt]
	\end{tabular}
}
\end{table}

% ------------- Results on Human3.6M (with occlusion) --------------
\begin{table}[!t]
	\centering
	\caption{Comparison results on Human3.6M \cite{ionescu--tpami2014-human3.6m} with occlusion.}
	\label{tbl:occlusion_human}
	\small
	\setlength{\tabcolsep}{0.85mm}{
	\begin{tabular}{l c c l l l}
		\toprule[0.75pt]
		Method & Venue & Occ. & MSE$\downarrow$ & MAE$\downarrow$ & SSIM$\uparrow$ \\
		\midrule[0.5pt]
		\rowcolor{maroon}PhyDNet\cite{guen-cvpr2020-phydnet}       & CVPR'20    &            & 36.9       & 16.2       & 0.901         \\
		&            & \checkmark & 39.2(+2.3) & 17.9(+1.7) & 0.870 \\
		\rowcolor{maroon}MAU\cite{chang-nips2021-mau}              & NeurIPS'21 &            & 33.1       & 14.9       & 0.883         \\
		&            & \checkmark & 50.7(+17.6)& 22.5(+7.6) & 0.830 \\
		\rowcolor{maroon}SimVP\cite{gao-cvpr2022-simvp}            & CVPR'22    &            & 31.6       & 15.1       & 0.904         \\
		&            & \checkmark & 33.4(+1.8) & 16.4(+1.3) & 0.897 \\
		\rowcolor{maroon}\scriptsize{PredRNNv2}\cite{wang-tpami2023-predrnnv2}  & TPAMI'23   &            & 34.8       & 17.2       & 0.864         \\
		&            & \checkmark & 36.8(+2.0) & 18.7(+1.5) & 0.842 \\
		\midrule[0.5pt]
		\rowcolor{maroon}Ours                                      &            &            & 23.3       & 11.9       & 0.912         \\
		Ours  &  w/o Inpainter   & \checkmark & \underline{24.7}(+1.4) & \underline{13.0}(+1.1) & \underline{0.906} \\
		Ours   & w/ Inpainter  & \checkmark & \textbf{24.4}(+1.1)    & \textbf{12.8}(+0.9)    & \textbf{0.907}    \\
		\toprule[0.75pt]
	\end{tabular}
}
\end{table}

% ------------- Results on Caltech Pedestrian (with occlusion) --------------
\begin{table}[!t]
	\centering
	\caption{Comparison results on Caltech Pedestrian \cite{dollar-cvpr2009-caltech} with occlusion.}
	\label{tbl:occlusion_caltech}
	\small
	\setlength{\tabcolsep}{0.85mm}{
		\begin{tabular}{l c c l l l}
			\toprule[0.75pt]
			Method & Venue & Occ. & MSE$\downarrow$ & SSIM$\uparrow$ & PSNR$\uparrow$(db) \\
			\midrule[0.5pt]
			\rowcolor{maroon}CrevNet\cite{yu-iclr2020-crevnet}    & ICLR'20    &            & 1.55                    & 0.925         & 29.3       \\
			&            & \checkmark & 2.86(+1.31)             & 0.878  & 24.7(-4.6) \\
			\rowcolor{maroon}\scriptsize{STMFANet}\cite{jin-cvpr2020-stmfanet} & CVPR'20    &            & 1.59                    & 0.927         & 29.1       \\
			&            & \checkmark & 3.18(+1.59)             & 0.874  & 25.6(-5.5) \\
			\rowcolor{maroon}MAU\cite{chang-nips2021-mau}         & NeurIPS'21 &            & 1.24                    & 0.943         & 30.1       \\
			&            & \checkmark & 2.55(+1.31)             & 0.898  & 24.6(-5.5) \\
			\rowcolor{maroon}SimVP\cite{gao-cvpr2022-simvp}       & CVPR'22    &            & 1.59                    & 0.927         & 33.1       \\
			&            & \checkmark & 3.23(+1.67)             & 0.892  &30.1(-3.0)  \\
			\midrule[0.5pt]
			\rowcolor{maroon}Ours                                 &            &            & 1.14                    & 0.949         & 33.1       \\
			Ours        & w/o Inpainter    & \checkmark & \underline{2.19}(+1.05) & \underline{0.917}  & \underline{31.4}(-1.7) \\
			Ours         & w/ Inpainter     & \checkmark & \textbf{2.08}(+0.94) & \textbf{0.921}  & \textbf{32.2}(-0.9) \\
			
			\toprule[0.75pt]
		\end{tabular}
	}
\end{table}

% ------------- Results on KTH (with occlusion) --------------
\begin{table}[!t]
	\centering   
	\caption{Comparison results on KTH \cite{schuldt-icpr2004-kth} with occlusion.}
	\label{tbl:occlusion_kth}
	\small
	\setlength{\tabcolsep}{0.95mm}{
	\begin{tabular}{l c c l l }
		\toprule[0.75pt]
		Method & Venue & Occ. & SSIM$\uparrow$ & PSNR$\uparrow$(db) \\
		\midrule[0.5pt]
		\rowcolor{maroon}\scriptsize{STMFANet}\cite{jin-cvpr2020-stmfanet}     & CVPR'20 &            & 0.893         & 29.85         \\
		&         & \checkmark & 0.871(-0.022) & 26.27(-3.58)  \\
		\rowcolor{maroon}LMC\cite{lee-cvpr2021-lmc}               & CVPR'21 &            & 0.894         & 28.61         \\
		&         & \checkmark & 0.879(-0.015) & 26.28(-2.33)  \\
		\rowcolor{maroon}SimVP\cite{gao-cvpr2022-simvp}           & CVPR'22 &            & 0.905         & 33.72         \\
		&         & \checkmark & 0.895(-0.010) & 33.00(-0.72)  \\
		\rowcolor{maroon}\scriptsize{PredRNNv2}\cite{wang-tpami2023-predrnnv2} & TPAMI'23&            & 0.858         & 29.79         \\
		&         & \checkmark & 0.827(-0.031) & 25.39(-4.40)  \\
		\midrule[0.5pt]
		\rowcolor{maroon}Ours                                     &         &            & 0.912         & 34.24         \\
		Ours   &  w/o Inpainter    & \checkmark & \underline{0.904}(-0.008) & \underline{33.57}(-0.67) \\
		Ours      & w/ Inpainter     & \checkmark & \textbf{0.905}(-0.007)    & \textbf{33.69}(-0.55)    \\
		\toprule[0.75pt]
	\end{tabular}
}
\end{table}

\subsection{Ablation Study}
We conduct the ablations on the FFT Inception block with different Fourier Units (without inpainter), and the recovery loss with different hyper-parameters $\lambda$ (with inpainter) on the test data. Note that it requires two days to train for 2000 epochs until convergence on Moving MNIST, which is very long, so we run 100 epochs in the ablations to save time. The parameters keep the same as in training unless specified.

\begin{figure}[!t]
	\centering
	\subfigure[]{\includegraphics[width=0.49\linewidth]{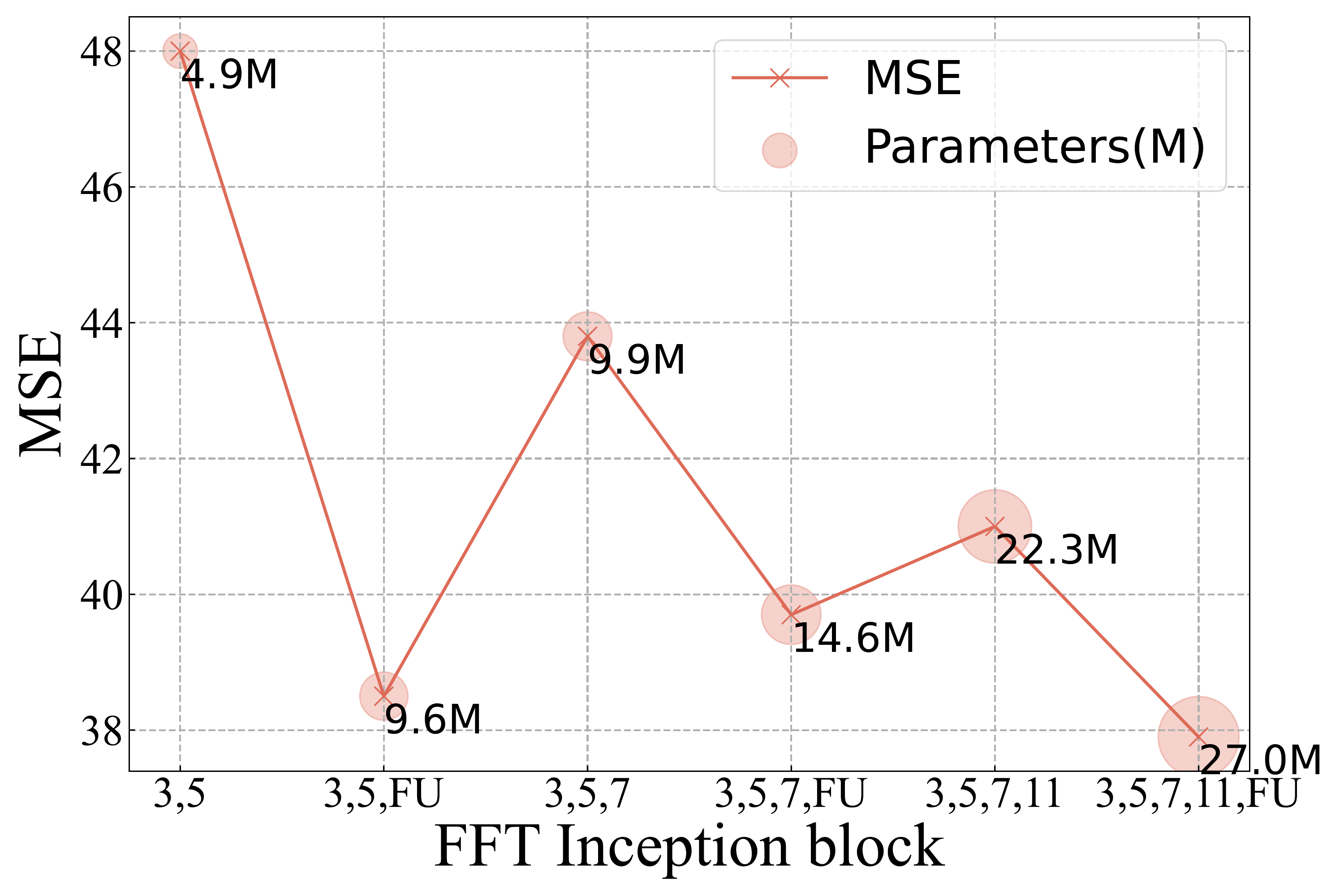}}
	\subfigure[]{\includegraphics[width=0.49\linewidth]{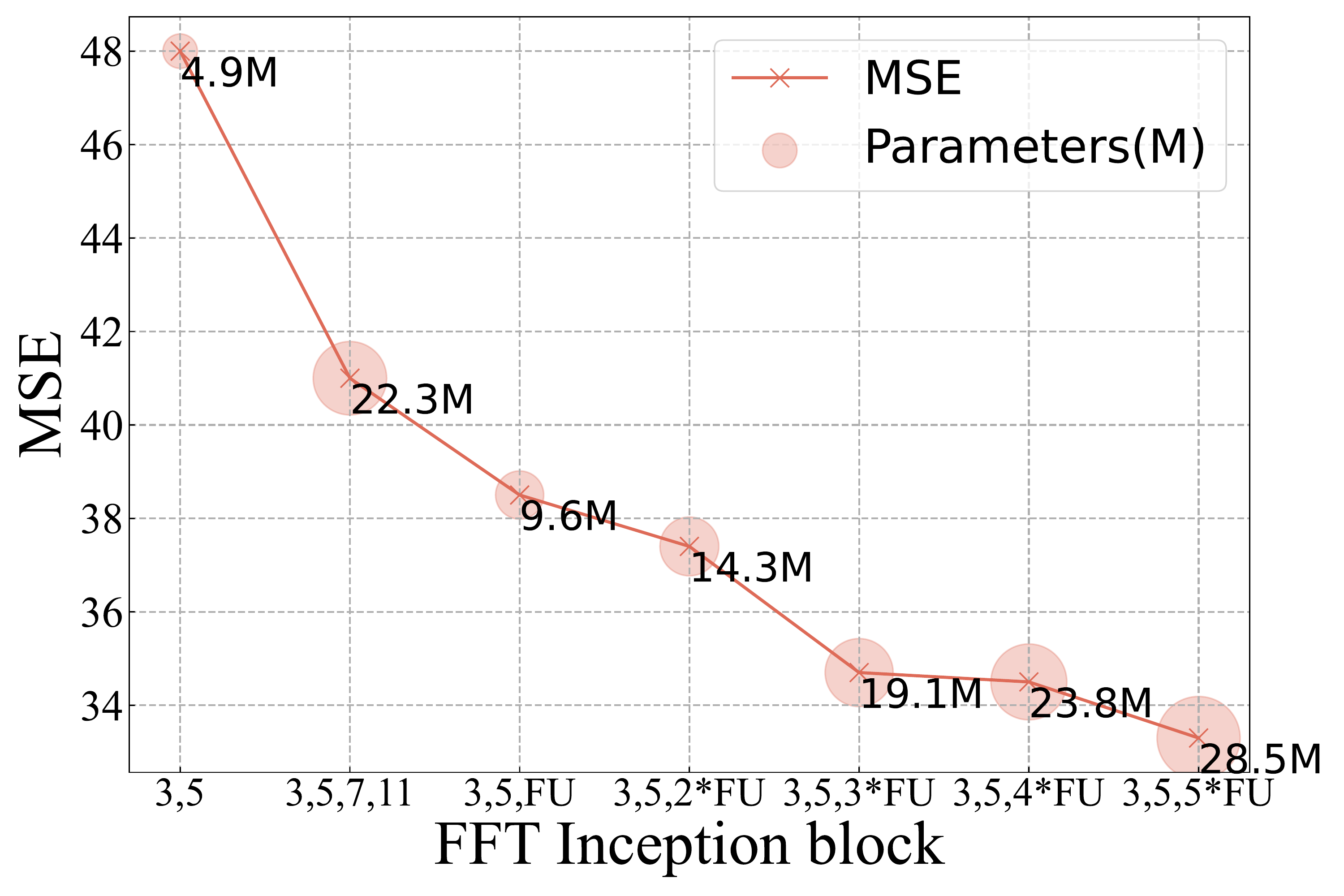}}
	\caption{Ablations of the FFT Inception block on Moving MNIST \cite{srivastava-icml2015-movingmnist}. (a) Different Fourier Unit (FU) positions; (b) Varying numbers of FUs. We use the comma to separate different branches.}
	\label{fig:ablation_fu_mnist}
\end{figure}

\textbf{FFT Inception block}. We explore the performance of our model using different architectures in the FFT Inception block on Moving MNIST \cite{srivastava-icml2015-movingmnist}. In particular, we vary the group convolution kernel size from 3 to 11, and insert the Fourier Unit (FU) after the second, the third, and the fourth group convolution branch; the results are shown in Fig.~\ref{fig:ablation_fu_mnist}(a). From the left figure, we observe that when using more group convolution branches, the performance is unnecessarily getting better. On the contrary, it gets a good trade-off between the performance and the model size, when using two group convolutions followed by the Fourier Unit. Moreover, we vary the number of FUs from 1 to 5, and depict the results in Fig.~\ref{fig:ablation_fu_mnist}(b). From the right figure, it can be seen that the prediction performance becomes better when using more FUs, but the model size becomes much larger. Overall, it seems that the performance gets a good balance with two group convolutions and three FUs. 

In addition, we vary the number of FUs $M$ from 1 to 3 on the other datasets and show the results in Table~\ref{tbl:ablation_M}. From the table, we see that our FFINet model achieves the best when using two FUs on TaxiBJ \cite{zhang-aaai2017-trafficbj}, Human3.6M \cite{ionescu--tpami2014-human3.6m}, and Caltech Pedestrian \cite{geiger-ijrr2013-kitti}, and one FU on KTH \cite{schuldt-icpr2004-kth}. It suggests a modest number of FUs is enough to achieve promising prediction performance.

% ------------- Ablation of FFT Inception block on Moving MNIST ------------

% ------------- Ablation on the number (M) of Fourier Units (with occlusion) --------------
\begin{table}[!t]
	\centering
	\caption{MSE of our model with different $M$ Fourier Units.  }
	\label{tbl:ablation_M}
	\begin{tabular}{l c c c }
		\toprule[0.75pt]
		Dataset & $M =1$ & $M =2$ & $M =3$ \\
		\midrule[0.5pt]
		MovingMNIST\cite{srivastava-icml2015-movingmnist} & 38.5      & 37.4          & \textbf{34.7}\\   
		TaxiBJ\cite{zhang-aaai2017-trafficbj}          & 43.3          & \textbf{41.2} & 43.4\\
		Human3.6M\cite{ionescu--tpami2014-human3.6m}   & 23.8          & \textbf{23.3} & 24.1 \\
		CaltechPed.\cite{geiger-ijrr2013-kitti}            & 1.16          & \textbf{1.14} & 1.21 \\
		KTH\cite{schuldt-icpr2004-kth}                 & \textbf{25.4} & 26.3          & 28.8 \\
		\toprule[0.75pt]
	\end{tabular}
\end{table}

% --------- Ablations on the recovery loss with different $\lambda$ ---------
\begin{table*}[!t]
	\centering
	\caption{Ablations on the recovery loss with different $\lambda$.}
	\label{tbl:ablation_recoverloss}
	\small
	\setlength{\tabcolsep}{1.1mm}{
		\begin{tabular}{c ccc c ccc c ccc c ccc c ccc}
			\toprule[0.75pt]
			\multirow{2}{*}{$\lambda$} & \multicolumn{4}{c}{Moving MNIST\cite{srivastava-icml2015-movingmnist}} & \multicolumn{4}{c}{TaxiBJ\cite{zhang-aaai2017-trafficbj}} & \multicolumn{4}{c}{Human3.6M\cite{ionescu--tpami2014-human3.6m}} & \multicolumn{4}{c}{Caltech Pedestrian\cite{dollar-cvpr2009-caltech}}  & \multicolumn{3}{c}{KTH\cite{schuldt-icpr2004-kth}}\\ \cmidrule{2-4}  \cmidrule{6-8} \cmidrule{10-12} \cmidrule{14-16} \cmidrule{18-20} 
			& MSE$\downarrow$ & MAE$\downarrow$ & SSIM$\uparrow$ & & MSE$\downarrow$ & MAE$\downarrow$ & SSIM$\uparrow$ & & MSE$\downarrow$ & MAE$\downarrow$ & SSIM$\uparrow$ && MSE$\downarrow$ & MAE$\downarrow$ &SSIM$\uparrow$ && MSE$\downarrow$ & MAE$\downarrow$ &SSIM$\uparrow$ \\ \midrule[0.5pt]
			0.00    & 42.8 & 114.6 & 0.898 & & 40.6 & 16.1 & 0.983 && 25.2 & 13.5 & 0.905 && 2.4 & 19.1 & 0.919  && 42.6 & 418.5& 0.899  \\
			0.25	& 41.4 & 110.3 & 0.901 & & 40.6 & 16.1 & 0.983 && 24.9 & 13.3 & 0.905 && 2.3 & 19.1 & 0.919  && 42.4 & 416.4& 0.901  \\
			0.50    & \textbf{40.3}& \textbf{105.5} & \textbf{0.908} & & 40.5 & 16.1 & 0.983 && 25.0 & 13.0 & 0.905 && 2.2 & 18.9 & 0.920  && 41.8 & 415.9&0.903 \\
			1.00    & 42.2 & 111.0 & 0.900 & & \textbf{40.4} & \textbf{16.0} & \textbf{0.983} && \textbf{24.4}& \textbf{12.8}& \textbf{0.907}&& \textbf{2.1} & \textbf{18.8} & \textbf{0.921} && \textbf{40.5} & \textbf{413.7} & \textbf{0.905} \\
			2.00    & 42.6 & 112.7 & 0.896 & & 40.6 & 16.1 & 0.983 && 24.6 & 13.0 & 0.905 && 2.4 & 19.2 & 0.919  && 41.9 & 416.2 & \textbf{0.905}  \\
			\toprule[0.75pt]
		\end{tabular}
	}
\end{table*}

\textbf{Recovery loss}. The recovery loss estimates the error between the source frame and the recovery frame, and its contribution to the model is governed by the hyper-parameter $\lambda$. We vary its value from 0 to 2, and show the results in Table~\ref{tbl:ablation_recoverloss}. From the table, we observe that our approach improves the video prediction performance with the recovery loss across all datasets. This demonstrates that the inpainting quality of the frames directly influences the future frame prediction. In particular, it achieves the best performance when $\lambda$ is set to 0.5 for Moving MNIST \cite{srivastava-icml2015-movingmnist} and 1.0 for the rest. 

\subsection{Qualitative Results}
\begin{figure}[!t]
	\centering
	\includegraphics[width=0.48\textwidth]{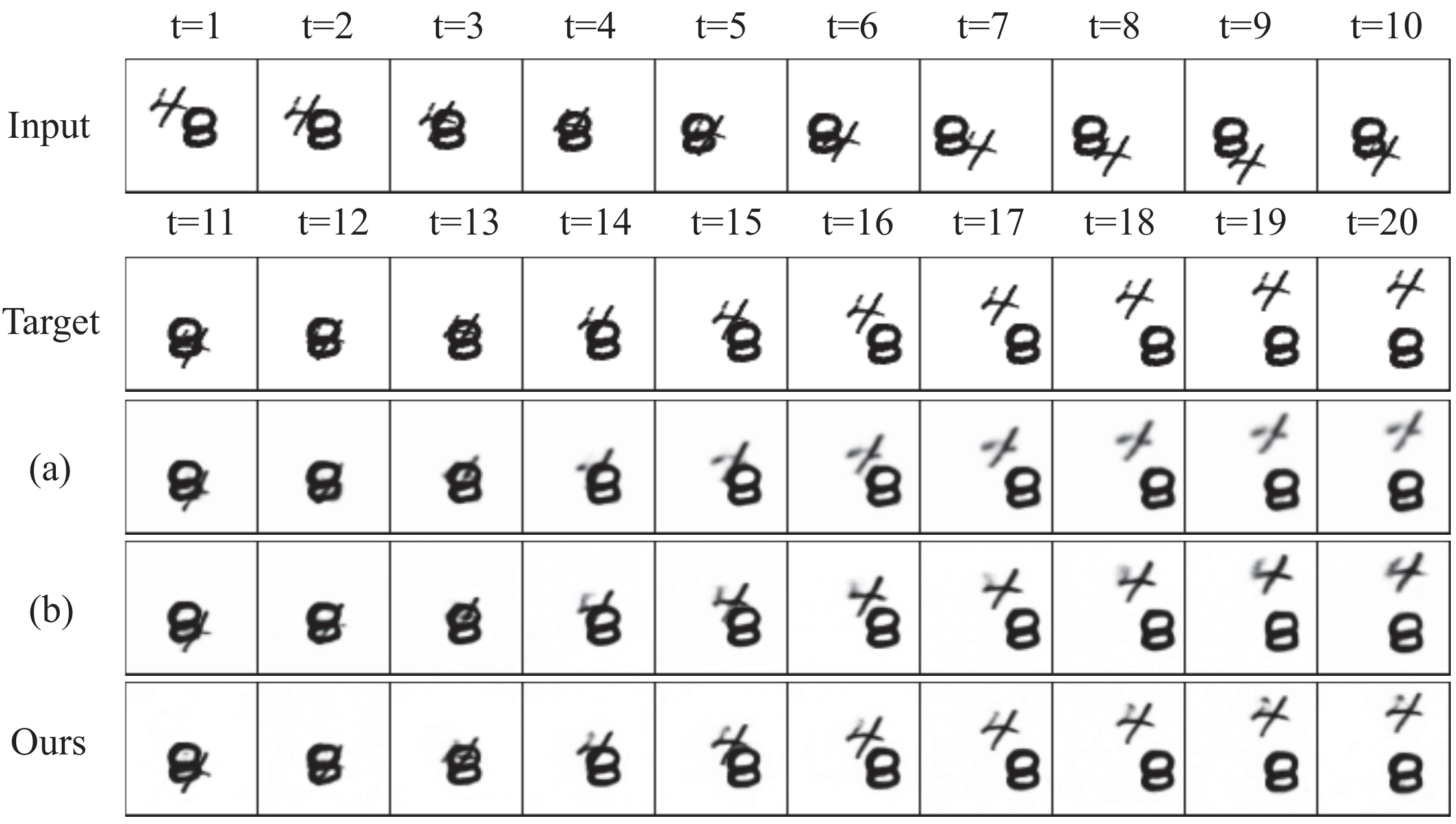}
	\includegraphics[width=0.48\textwidth]{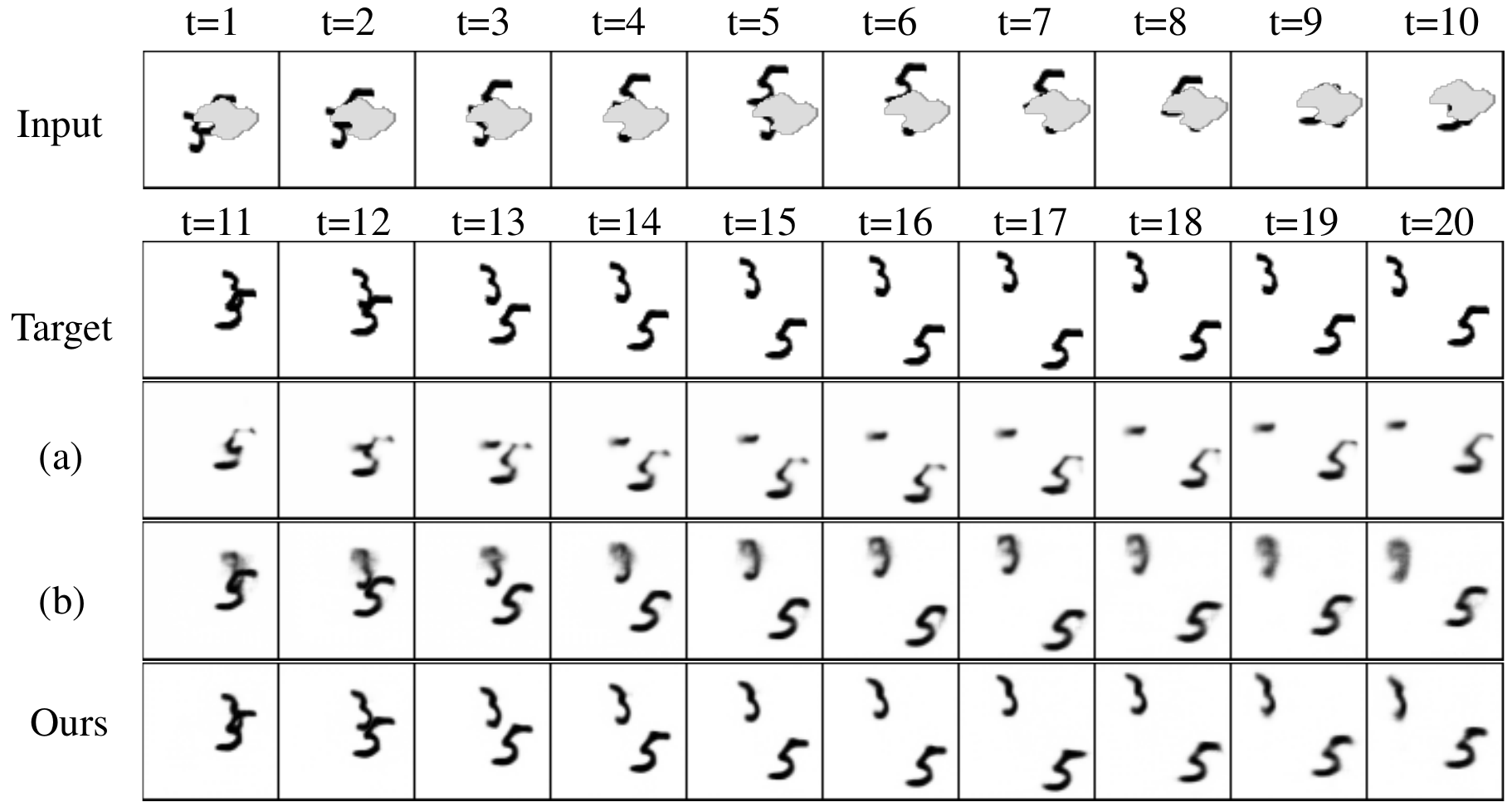}
	\caption{Predictions on Moving MNIST. (a) PhyDNet \cite{guen-cvpr2020-phydnet}; (b) SimVP \cite{gao-cvpr2022-simvp}.}
	\label{fig:mnist}
\end{figure}

\begin{figure}[!t]
	\centering
	\includegraphics[width=0.48\textwidth]{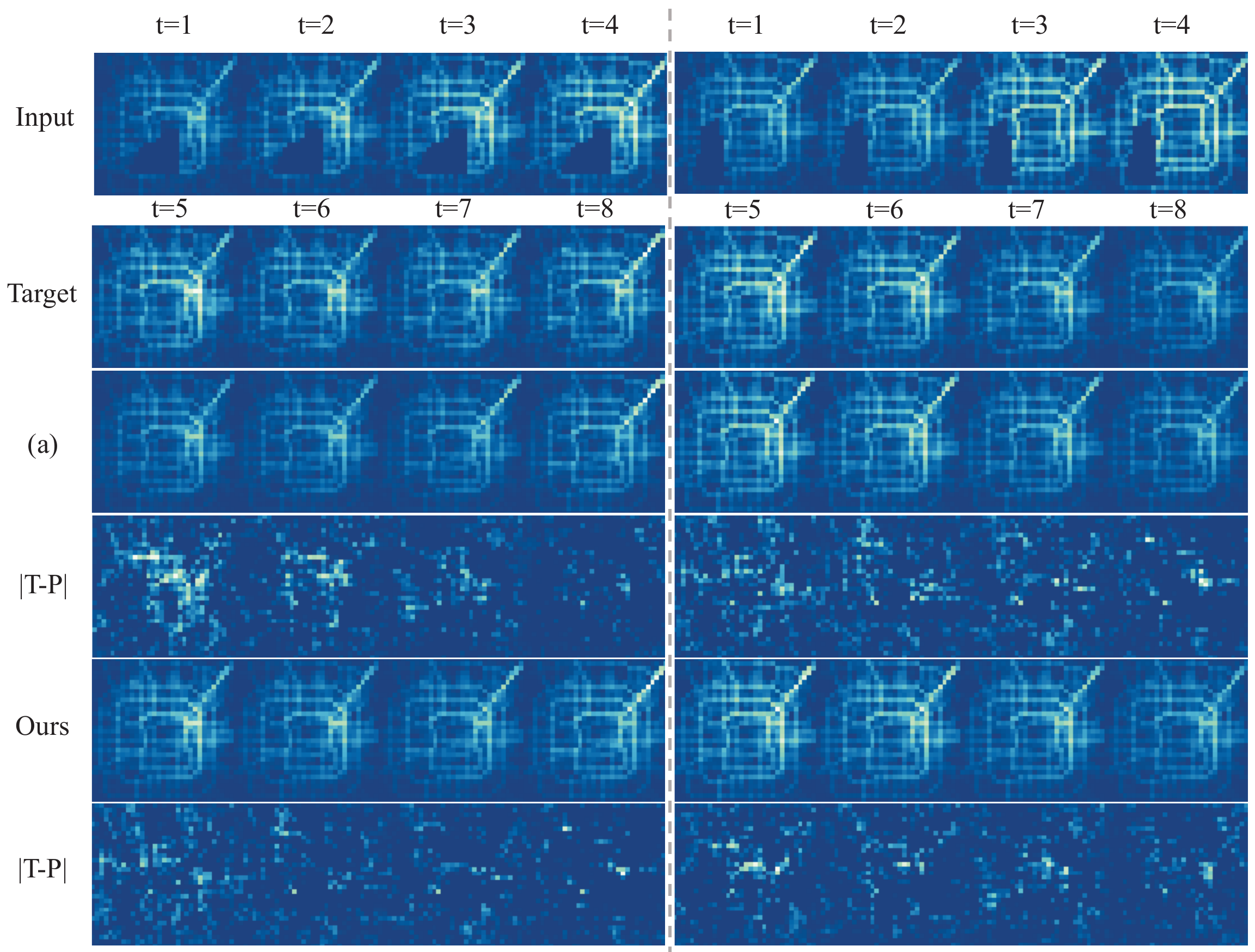}
	\caption{Predictions on TaxiBJ \cite{zhang-aaai2017-trafficbj}. (a) PredRNNv2 \cite{wang-tpami2023-predrnnv2}.}
	\label{fig:taxibj}
\end{figure}

\begin{figure}[!t]
	\centering
	\includegraphics[width=0.48\textwidth]{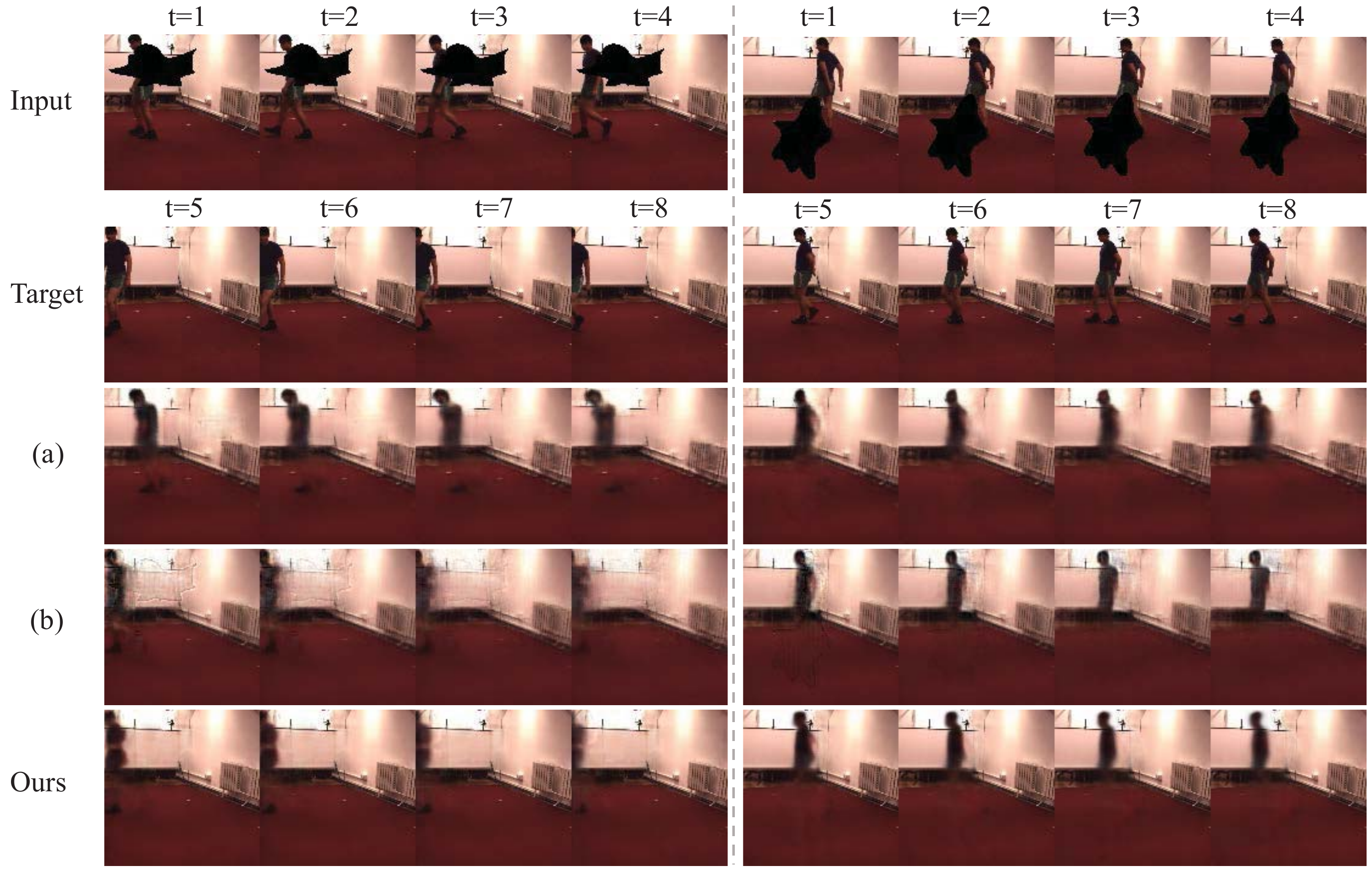}
	\caption{Predictions on Human3.6M \cite{ionescu--tpami2014-human3.6m}. (a) MAU \cite{chang-nips2021-mau}; (b) SimVP \cite{gao-cvpr2022-simvp}.}
	\label{fig:human3.6m}
\end{figure}

\begin{figure}[!t]
	\centering
	\includegraphics[width=0.45\textwidth]{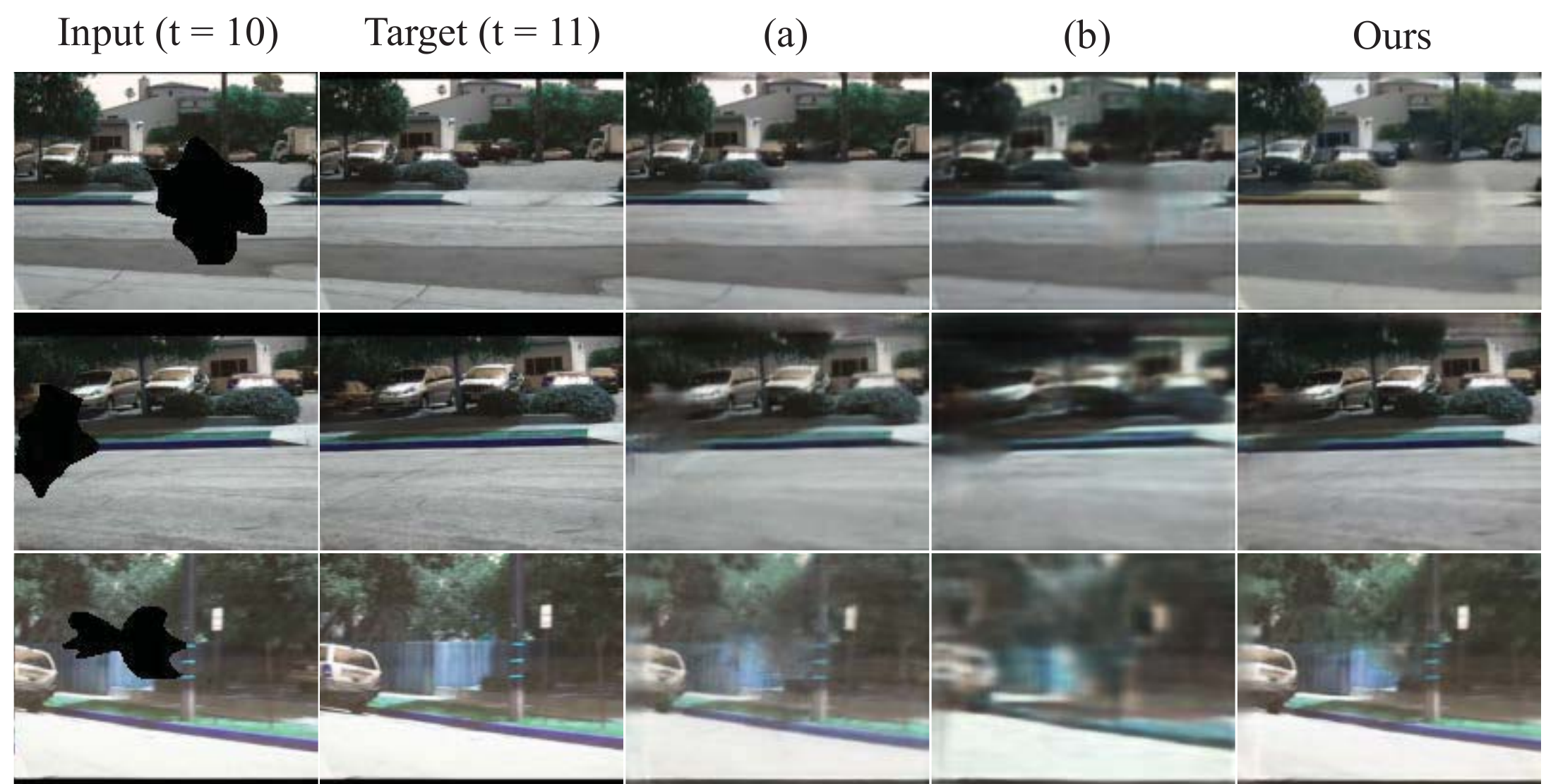}
	\caption{Predictions on Caltech Ped. \cite{dollar-cvpr2009-caltech}. (a) CrevNet \cite{yu-iclr2020-crevnet}; (b) SimVP \cite{gao-cvpr2022-simvp}.}
	\label{fig:caltech}
\end{figure}

\begin{figure}[!t]
	\centering
	\includegraphics[width=0.48\textwidth]{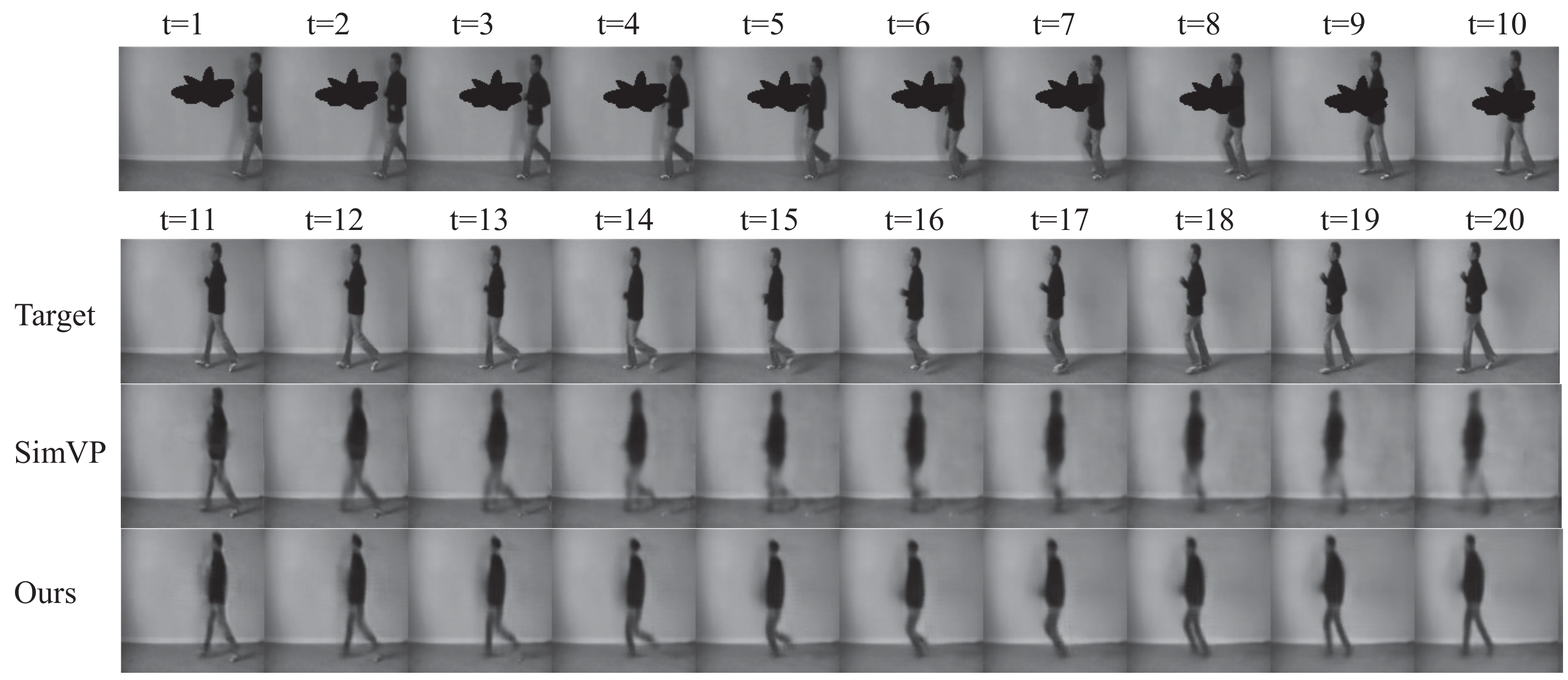}
	\caption{Predictions on KTH \cite{schuldt-icpr2004-kth}.}
	\label{fig:kth}
\end{figure}

To give an intuitive view on the superiority of our model, we select some challenging cases with occlusion from the five datasets and visualize the predicted frames in Fig.~\ref{fig:mnist} to Fig.~\ref{fig:kth}. Besides, we show the predicted frames without occlusion on Moving MNIST in Fig.~\ref{fig:mnist}, where the two digits are seriously overlapped. 

In Fig.~\ref{fig:mnist}, the two digits can be well generated by our method even when the input digits are heavily occluded. This is because the designed inpainter is able to fill in the missing area of the frame before the translator to capture the temporal dynamics in video. On the contrary, previous methods like PhyDNet \cite{guen-cvpr2020-phydnet} and SimVP \cite{gao-cvpr2022-simvp} learns the spatiotemporal features of frames with occlusion, which might bring about some misleading information, leading to inferior predictions.  

In Fig.~\ref{fig:taxibj}, the difference map $|T-P|$ of the target frame and the predicted frame reflects the quality of the traffic flow prediction. When the dark area becomes larger, it means the prediction is more accurate. From the figure, we see that the dark area produced by our model is much larger that the compared PredRNNv2 \cite{wang-tpami2023-predrnnv2} that does not consider the occlusion scenario.

Fig.~\ref{fig:human3.6m} shows a man walking through a room and Fig.~\ref{fig:caltech} shows the street scenario. Here, the masks are randomly placed in different positions. From the figures, we observe that our method generates higher-quality frames compared to others, \eg, the human body and the car are more clearly to see. Fig.~\ref{fig:kth} show the walking action clip with the occlusion in the middle, and we see that the predicted sequence by ours matches better with the target ones. This is because the developed inpainter is able to recover the occluded area by adopting the fast Fourier convolution, and the designed translator models the temporal dynamics by capturing both the local and the global spatiotemporal features.

% ------------- Visualization on Moving MNIST -------

% ------------- Visualization on TaxiBJ -------

% ------------- Visualization on Human3.6M -------

% ------------- Visualization on Caltech Pedestrian -------

% ------------- Visualization on KTH -------

%------------------------------------------------------------------------
\section{Conclusion}
\label{conclusion}
In this work, we have explored the occluded video prediction problem, which appears frequently in practice but remains untouched yet. To address the occlusion issue, we design the inpainter module which employs the channel-wise fast Fourier convolution to enlarge the receptive field, thus capturing the global context to recover the missing area in the frame. To model the temporal dynamics, we develop the Fast Fourier Transform Inception block that includes group convolutions and multiple Fourier Units to learn both the local and the global spatiotemporal features, which help to capture the temporal evolution across the video frames. Hence, we proposed the fully-convolutional Fast Fourier Inception Networks, terms FFINet, for occluded video prediction, and conducted comprehensive experiments to verify the effectiveness of the proposed method on several benchmarks.
%

%\bibliographystyle{IEEEtran}
%\bibliography{videopre}

\begin{thebibliography}{10}
	\providecommand{\url}[1]{#1}
	\csname url@samestyle\endcsname
	\providecommand{\newblock}{\relax}
	\providecommand{\bibinfo}[2]{#2}
	\providecommand{\BIBentrySTDinterwordspacing}{\spaceskip=0pt\relax}
	\providecommand{\BIBentryALTinterwordstretchfactor}{4}
	\providecommand{\BIBentryALTinterwordspacing}{\spaceskip=\fontdimen2\font plus
		\BIBentryALTinterwordstretchfactor\fontdimen3\font minus
		\fontdimen4\font\relax}
	\providecommand{\BIBforeignlanguage}[2]{{%
			\expandafter\ifx\csname l@#1\endcsname\relax
			\typeout{** WARNING: IEEEtran.bst: No hyphenation pattern has been}%
			\typeout{** loaded for the language `#1'. Using the pattern for}%
			\typeout{** the default language instead.}%
			\else
			\language=\csname l@#1\endcsname
			\fi
			#2}}
	\providecommand{\BIBdecl}{\relax}
	\BIBdecl
	
	\bibitem{shi-nips2015-convlstm}
	X.~Shi, Z.~Chen, H.~Wang, D.~Yeung, W.~Wong, and W.~Woo, ``Convolutional {LSTM}
	network: {A} machine learning approach for precipitation nowcasting,'' in
	\emph{Advances in Neural Information Processing Systems (NeurIPS)}, 2015, pp.
	802--810.
	
	\bibitem{wu-cvpr2021-motionrnn}
	H.~Wu, Z.~Yao, J.~Wang, and M.~Long, ``Motionrnn: {A} flexible model for video
	prediction with spacetime-varying motions,'' in \emph{Proceedings of the IEEE
		Conference on Computer Vision and Pattern Recognition (CVPR)}, 2021, pp.
	15\,435--15\,444.
	
	\bibitem{bei-cvpr2021-sadm}
	X.~Bei, Y.~Yang, and S.~Soatto, ``Learning semantic-aware dynamics for video
	prediction,'' in \emph{Proceedings of the IEEE Conference on Computer Vision
		and Pattern Recognition (CVPR)}, 2021, pp. 902--912.
	
	\bibitem{castrejon-iccv2019-conditionalvrnn}
	L.~Castrej{\'{o}}n, N.~Ballas, and A.~C. Courville, ``Improved conditional
	vrnns for video prediction,'' in \emph{Proceedings of the IEEE International
		Conference on Computer Vision (ICCV)}, 2019, pp. 7607--7616.
	
	\bibitem{babaeizadeh-iclr2018-sv2p}
	M.~Babaeizadeh, C.~Finn, D.~Erhan, R.~H. Campbell, and S.~Levine, ``Stochastic
	variational video prediction,'' in \emph{Proceedings of the International
		Conference on Learning Representations (ICLR)}, 2018.
	
	\bibitem{denton-icml2018-svg}
	E.~Denton and R.~Fergus, ``Stochastic video generation with a learned prior,''
	in \emph{Proceedings of the International Conference on Machine Learning
		(ICML)}, 2018, pp. 1182--1191.
	
	\bibitem{elman-cognsci1990-rnn}
	J.~L. Elman, ``Finding structure in time,'' \emph{Cognitive Science}, vol.~14,
	no.~2, pp. 179--211, 1990.
	
	\bibitem{lecun-proc1998-cnn}
	Y.~LeCun, L.~Bottou, Y.~Bengio, and P.~Haffner, ``Gradient-based learning
	applied to document recognition,'' \emph{Proc. {IEEE}}, vol.~86, no.~11, pp.
	2278--2324, 1998.
	
	\bibitem{wang-nips2017-predrnn}
	Y.~Wang, M.~Long, J.~Wang, Z.~Gao, and P.~S. Yu, ``Predrnn: Recurrent neural
	networks for predictive learning using spatiotemporal lstms,'' in
	\emph{Advances in Neural Information Processing Systems (NeurIPS)}, 2017, pp.
	879--888.
	
	\bibitem{wang-iclr2019-e3dlstm}
	Y.~Wang, L.~Jiang, M.~Yang, L.~Li, M.~Long, and L.~Fei{-}Fei, ``Eidetic 3d
	{LSTM:} {A} model for video prediction and beyond,'' in \emph{Proceedings of
		the International Conference on Learning Representations (ICLR)}, 2019.
	
	\bibitem{chang-nips2021-mau}
	Z.~Chang, X.~Zhang, S.~Wang, S.~Ma, Y.~Ye, X.~Xiang, and W.~Gao, ``Mau: A
	motion-aware unit for video prediction and beyond,'' in \emph{Advances in
		Neural Information Processing Systems (NeurIPS)}, vol.~34, 2021.
	
	\bibitem{liu-iccv2017-dvf}
	Z.~Liu, R.~A. Yeh, X.~Tang, Y.~Liu, and A.~Agarwala, ``Video frame synthesis
	using deep voxel flow,'' in \emph{Proceedings of the IEEE International
		Conference on Computer Vision (ICCV)}, 2017, pp. 4473--4481.
	
	\bibitem{gao-iccv2019-dpg}
	H.~Gao, H.~Xu, Q.~Cai, R.~Wang, F.~Yu, and T.~Darrell, ``Disentangling
	propagation and generation for video prediction,'' in \emph{Proceedings of
		the IEEE International Conference on Computer Vision (ICCV)}, 2019, pp.
	9005--9014.
	
	\bibitem{gao-cvpr2022-simvp}
	Z.~Gao, C.~Tan, L.~Wu, and S.~Z. Li, ``Simvp: Simpler yet better video
	prediction,'' in \emph{Proceedings of the IEEE Conference on Computer Vision
		and Pattern Recognition (CVPR)}, 2022, pp. 3160--3170.
	
	\bibitem{dosovitskiy-iclr2021-vit}
	A.~Dosovitskiy, L.~Beyer, A.~Kolesnikov, D.~Weissenborn, X.~Zhai,
	T.~Unterthiner, M.~Dehghani, M.~Minderer, G.~Heigold, S.~Gelly, J.~Uszkoreit,
	and N.~Houlsby, ``An image is worth 16x16 words: transformers for image
	recognition at scale,'' in \emph{Proceedings of the International Conference
		on Learning Representations (ICLR)}, 2021.
	
	\bibitem{yu-iclr2020-crevnet}
	W.~Yu, Y.~Lu, S.~Easterbrook, and S.~Fidler, ``Efficient and
	information-preserving future frame prediction and beyond,'' in
	\emph{Proceedings of the International Conference on Learning Representations
		(ICLR)}, 2020.
	
	\bibitem{wang-tmm2023-stsanet}
	Z.~Wang, Z.~Liu, G.~Li, Y.~Wang, T.~Zhang, L.~Xu, and J.~Wang,
	``Spatio-temporal self-attention network for video saliency prediction,''
	\emph{IEEE Transactions on Multimedia (TMM)}, vol.~25, pp. 1161--1174, 2023.
	
	\bibitem{chi-nips2020-ffc}
	L.~Chi, B.~Jiang, and Y.~Mu, ``Fast fourier convolution,'' in \emph{Advances in
		Neural Information Processing Systems (NeurIPS)}, 2020.
	
	\bibitem{katznelson-amm2005-harmonic}
	Y.~Katznelson, ``An introduction to harmonic analysis,'' \emph{The American
		Mathematical Monthly}, vol.~77, no.~4, 2005.
	
	\bibitem{suvorov-wacv2022-lama}
	R.~Suvorov, E.~Logacheva, A.~Mashikhin, A.~Remizova, A.~Ashukha, A.~Silvestrov,
	N.~Kong, H.~Goka, K.~Park, and V.~Lempitsky, ``Resolution-robust large mask
	inpainting with fourier convolutions,'' in \emph{Proceedings of the IEEE
		Winter Conference on Applications of Computer Vision (WACV)}, 2022, pp.
	3172--3182.
	
	\bibitem{srivastava-icml2015-movingmnist}
	N.~Srivastava, E.~Mansimov, and R.~Salakhutdinov, ``Unsupervised learning of
	video representations using lstms,'' in \emph{Proceedings of the
		International Conference on Machine Learning (ICML)}, 2015, pp. 843--852.
	
	\bibitem{zhang-aaai2017-trafficbj}
	J.~Zhang, Y.~Zheng, and D.~Qi, ``Deep spatio-temporal residual networks for
	citywide crowd flows prediction,'' in \emph{Proceedings of the AAAI
		conference on artificial intelligence(AAAI)}, 2017, pp. 1655--1661.
	
	\bibitem{ionescu--tpami2014-human3.6m}
	C.~Ionescu, D.~Papava, V.~Olaru, and C.~Sminchisescu, ``Human3.6m: Large scale
	datasets and predictive methods for 3d human sensing in natural
	environments,'' \emph{IEEE Transactions on Pattern Analysis and Machine
		Intelligence (TPAMI)}, vol.~36, no.~7, pp. 1325--1339, 2014.
	
	\bibitem{geiger-ijrr2013-kitti}
	A.~Geiger, P.~Lenz, C.~Stiller, and R.~Urtasun, ``Vision meets robotics: The
	{KITTI} dataset,'' \emph{International Journal of Robotics Research (IJRR)},
	vol.~32, no.~11, pp. 1231--1237, 2013.
	
	\bibitem{schuldt-icpr2004-kth}
	C.~Sch{\"{u}}ldt, I.~Laptev, and B.~Caputo, ``Recognizing human actions: {A}
	local {SVM} approach,'' in \emph{Proceedings of the Computational Vision and
		Active Perception Laboratory (CVAP)}, 2004, pp. 32--36.
	
	\bibitem{wang-icml2018-predrnn++}
	Y.~Wang, Z.~Gao, M.~Long, J.~Wang, and P.~S. Yu, ``Predrnn++: Towards {a}
	resolution of the deep-in-time dilemma in spatiotemporal predictive
	learning,'' in \emph{Proceedings of the International Conference on Machine
		Learning (ICML)}, 2018, pp. 5110--5119.
	
	\bibitem{wang-cvpr2019-mim}
	Y.~Wang, J.~Zhang, H.~Zhu, M.~Long, J.~Wang, and P.~S. Yu, ``Memory in memory:
	{A} predictive neural network for learning higher-order non-stationarity from
	spatiotemporal dynamics,'' in \emph{Proceedings of the IEEE Conference on
		Computer Vision and Pattern Recognition (CVPR)}, 2019, pp. 9154--9162.
	
	\bibitem{oliu-eccv2018-frnn}
	M.~Oliu, J.~Selva, and S.~Escalera, ``Folded recurrent neural networks for
	future video prediction,'' in \emph{Proceedings of the European Conference on
		Computer Vision (ECCV)}, vol. 11218, 2018, pp. 745--761.
	
	\bibitem{su-nips2020-convttlstm}
	J.~Su, W.~Byeon, J.~Kossaifi, F.~Huang, J.~Kautz, and A.~Anandkumar,
	``Convolutional tensor-train {LSTM} for spatio-temporal learning,'' in
	\emph{Advances in Neural Information Processing Systems (NeurIPS)}, 2020.
	
	\bibitem{wang-tpami2023-predrnnv2}
	Y.~Wang, H.~Wu, J.~Zhang, Z.~Gao, J.~Wang, P.~S. Yu, and M.~Long, ``Predrnn:
	{A} recurrent neural network for spatiotemporal predictive learning,''
	\emph{IEEE Transactions on Pattern Analysis and Machine Intelligence
		(TPAMI)}, vol.~45, no.~2, pp. 2208--2225, 2023.
	
	\bibitem{guen-cvpr2020-phydnet}
	V.~L. Guen and N.~Thome, ``Disentangling physical dynamics from unknown factors
	for unsupervised video prediction,'' in \emph{Proceedings of the IEEE
		Conference on Computer Vision and Pattern Recognition (CVPR)}, 2020, pp.
	11\,474--11\,484.
	
	\bibitem{park-aaai2021-vidode}
	S.~Park, K.~Kim, J.~Lee, J.~Choo, J.~Lee, S.~Kim, and E.~Choi, ``Vid-ode:
	Continuous-time video generation with neural ordinary differential
	equation,'' in \emph{Proceedings of the AAAI conference on artificial
		intelligence (AAAI)}, 2021, pp. 2412--2422.
	
	\bibitem{kim-tmm2021-dmee}
	N.~Kim and J.~Kang, ``Dynamic motion estimation and evolution video prediction
	network,'' \emph{IEEE Transactions on Multimedia (TMM)}, vol.~23, pp.
	3986--3998, 2021.
	
	\bibitem{ye-tmm2023-dck}
	Z.~Ye, M.~Xia, R.~Yi, J.~Zhang, Y.-K. Lai, X.~Huang, G.~Zhang, and Y.-J. Liu,
	``Audio-driven talking face video generation with dynamic convolution
	kernels,'' \emph{IEEE Transactions on Multimedia (TMM)}, vol.~25, pp.
	2033--2046, 2023.
	
	\bibitem{lee-cvpr2021-lmc}
	S.~Lee, H.~G. Kim, D.~H. Choi, H.~Kim, and Y.~M. Ro, ``Video prediction
	recalling long-term motion context via memory alignment learning,'' in
	\emph{Proceedings of the IEEE Conference on Computer Vision and Pattern
		Recognition (CVPR)}, 2021, pp. 3054--3063.
	
	\bibitem{chang-cvpr2022-strpm}
	Z.~Chang, X.~Zhang, S.~Wang, S.~Ma, and W.~Gao, ``{STRPM:} {A} spatiotemporal
	residual predictive model for high-resolution video prediction,'' in
	\emph{Proceedings of the IEEE Conference on Computer Vision and Pattern
		Recognition (CVPR)}, 2022, pp. 13\,926--13\,935.
	
	\bibitem{chen-cvpr2022-cpl}
	G.~Chen, W.~Zhang, H.~Lu, S.~Gao, Y.~Wang, M.~Long, and X.~Yang, ``Continual
	predictive learning from videos,'' in \emph{Proceedings of the IEEE
		Conference on Computer Vision and Pattern Recognition (CVPR)}, 2022, pp.
	10\,718--10\,727.
	
	\bibitem{yu-cvpr2022-mac}
	W.~Yu, W.~Chen, S.~Yin, S.~Easterbrook, and A.~Garg, ``Modular action concept
	grounding in semantic video prediction,'' in \emph{Proceedings of the IEEE
		Conference on Computer Vision and Pattern Recognition (CVPR)}, 2022, pp.
	3595--3604.
	
	\bibitem{weissenborn-iclr2020-subscalevitr}
	D.~Weissenborn, O.~T{\"{a}}ckstr{\"{o}}m, and J.~Uszkoreit, ``Scaling
	autoregressive video models,'' in \emph{Proceedings of the International
		Conference on Learning Representations (ICLR)}, 2020.
	
	\bibitem{rakhimov-visigrapp2021-latentvit}
	R.~Rakhimov, D.~Volkhonskiy, A.~Artemov, D.~Zorin, and E.~Burnaev, ``Latent
	video transformer,'' in \emph{Proceedings of the 16th International Joint
		Conference on Computer Vision, Imaging and Computer Graphics Theory and
		Applications}, 2021, pp. 101--112.
	
	\bibitem{yang-arxiv2021-tctn}
	Z.~Yang, X.~Yang, and Q.~Lin, ``Tctn: A 3d-temporal convolutional transformer
	network for spatiotemporal predictive learning,'' \emph{arXiv preprint
		arXiv:2112.01085}, 2021.
	
	\bibitem{wu-cvpr2022-ovp}
	Y.~Wu, Q.~Wen, and Q.~Chen, ``Optimizing video prediction via video frame
	interpolation,'' in \emph{Proceedings of the IEEE Conference on Computer
		Vision and Pattern Recognition (CVPR)}, 2022, pp. 17\,793--17\,802.
	
	\bibitem{huang-eccv2022-lcvp}
	J.~Huang, Y.~Jin, K.~M. Yi, and L.~Sigal, ``Layered controllable video
	generation,'' in \emph{Proceedings of the European Conference on Computer
		Vision (ECCV)}, vol. 13676, 2022, pp. 546--564.
	
	\bibitem{goodfellow-nips2014-gan}
	I.~Goodfellow, J.~Pouget-Abadie, M.~Mirza, B.~Xu, D.~Warde-Farley, S.~Ozair,
	A.~Courville, and Y.~Bengio, ``Generative adversarial nets,'' in
	\emph{Advances in Neural Information Processing Systems (NeurIPS)}, 2014, pp.
	1724--1734.
	
	\bibitem{kwon-cvpr2019-rcyclegan}
	Y.~Kwon and M.~Park, ``Predicting future frames using retrospective cycle
	{GAN},'' in \emph{Proceedings of the IEEE Conference on Computer Vision and
		Pattern Recognition (CVPR)}, 2019, pp. 1811--1820.
	
	\bibitem{xu-ijcv2021-pma}
	J.~Xu, B.~Ni, and X.~Yang, ``Progressive multi-granularity analysis for video
	prediction,'' \emph{International Journal of Computer Vision (IJCV)}, vol.
	129, no.~3, pp. 601--618, 2021.
	
	\bibitem{dollar-cvpr2009-caltech}
	P.~Doll{\'{a}}r, C.~Wojek, B.~Schiele, and P.~Perona, ``Pedestrian detection:
	{A} benchmark,'' in \emph{Proceedings of the IEEE Conference on Computer
		Vision and Pattern Recognition (CVPR)}, 2009, pp. 304--311.
	
	\bibitem{villegas-iclr2017-mcnet}
	R.~Villegas, J.~Yang, S.~Hong, X.~Lin, and H.~Lee, ``Decomposing motion and
	content for natural video sequence prediction,'' in \emph{Proceedings of the
		International Conference on Learning Representations (ICLR)}, 2017.
	
	\bibitem{wang-tip2004-ssim}
	Z.~Wang, A.~C. Bovik, H.~R. Sheikh, and E.~P. Simoncelli, ``Image quality
	assessment: from error visibility to structural similarity,'' \emph{{IEEE}
		Transactions on Image Processing (TIP)}, vol.~13, no.~4, pp. 600--612, 2004.
	
	\bibitem{chang-tmm2022-stam}
	Z.~Chang, X.~Zhang, S.~Wang, S.~Ma, and W.~Gao, ``Stam: A spatiotemporal
	attention based memory for video prediction,'' \emph{IEEE Transactions on
		Multimedia (TMM)}, vol.~25, pp. 2354--2367, 2023.
	
	\bibitem{kingma-iclr2015-adam}
	D.~P. Kingma and J.~Ba, ``Adam: {A} method for stochastic optimization,'' in
	\emph{Proceedings of the International Conference on Learning Representations
		(ICLR)}, 2015.
	
	\bibitem{simth-ispp2019-onecycle}
	L.~N.Smith and N.~Topin, ``Super-convergence: Very fast training of neural
	networks using large learning rates,'' in \emph{Proceedings of the
		International Society for Optics and Photonics}, 2019.
	
	\bibitem{li-cvpr2022-e2fgvi}
	Z.~Li, C.~Lu, J.~Qin, C.~Guo, and M.~Cheng, ``Towards an end-to-end framework
	for flow-guided video inpainting,'' in \emph{Proceedings of the IEEE
		Conference on Computer Vision and Pattern Recognition (CVPR)}, 2022, pp.
	17\,541--17\,550.
	
	\bibitem{liang-iccv2017-dualmotiongan}
	X.~Liang, L.~Lee, W.~Dai, and E.~P. Xing, ``Dual motion {GAN} for future-flow
	embedded video prediction,'' in \emph{Proceedings of the IEEE International
		Conference on Computer Vision (ICCV)}, 2017, pp. 1762--1770.
	
	\bibitem{lotter-iclr2017-prednet}
	W.~Lotter, G.~Kreiman, and D.~D. Cox, ``Deep predictive coding networks for
	video prediction and unsupervised learning,'' in \emph{Proceedings of the
		International Conference on Learning Representations (ICLR)}, 2017.
	
	\bibitem{hao-cvpr2018-ctrlgen}
	Z.~Hao, X.~Huang, and S.~J. Belongie, ``Controllable video generation with
	sparse trajectories,'' in \emph{Proceedings of the IEEE Conference on
		Computer Vision and Pattern Recognition (CVPR)}, 2018, pp. 7854--7863.
	
	\bibitem{byeon-eccv2018-contextvp}
	W.~Byeon, Q.~Wang, R.~K. Srivastava, and P.~Koumoutsakos, ``Contextvp: Fully
	context-aware video prediction,'' in \emph{Proceedings of the European
		Conference on Computer Vision (ECCV)}, 2018, pp. 781--797.
	
	\bibitem{jin-cvpr2020-stmfanet}
	B.~Jin, Y.~Hu, Q.~Tang, J.~Niu, Z.~Shi, Y.~Han, and X.~Li, ``Exploring
	spatial-temporal multi-frequency analysis for high-fidelity and
	temporal-consistency video prediction,'' in \emph{Proceedings of the IEEE
		Conference on Computer Vision and Pattern Recognition (CVPR)}, 2020, pp.
	4553--4562.
	
	\bibitem{geng-cvpr2022-vpcl}
	D.~Geng, M.~Hamilton, and A.~Owens, ``Comparing correspondences: Video
	prediction with correspondence-wise losses,'' in \emph{Proceedings of the
		IEEE Conference on Computer Vision and Pattern Recognition (CVPR)}, 2022, pp.
	3355--3366.
	
	\bibitem{jia-nips2016-dfn}
	X.~Jia, B.~D. Brabandere, T.~Tuytelaars, and L.~V. Gool, ``Dynamic filter
	networks,'' in \emph{Advances in Neural Information Processing Systems
		(NeurIPS)}, 2016, pp. 667--675.
	
	\bibitem{lee-iclr2019-savp}
	A.~X. Lee, R.~Zhang, F.~Ebert, P.~Abbeel, C.~Finn, and S.~Levine, ``Stochastic
	adversarial video prediction,'' in \emph{Proceedings of the 7th International
		Conference on Learning Representations (ICLR)}, 2019.
	
	\bibitem{gao-iros2021-gridkeypoint}
	X.~Gao, Y.~Jin, Q.~Dou, C.~Fu, and P.~Heng, ``Accurate grid keypoint learning
	for efficient video prediction,'' in \emph{Proceedings of the International
		Conference on Intelligent Robots and Systems (IROS)}, 2021, pp. 5908--5915.
	
\end{thebibliography}

\ifCLASSOPTIONcaptionsoff
  \newpage
\fi

\end{document}